\documentclass[nohyperref]{article}

\usepackage{microtype}
\usepackage{graphicx}
\usepackage{subfigure}
\usepackage{booktabs} 

\usepackage{hyperref}
\usepackage{arydshln}

\usepackage[accepted]{icml_preprint}

\usepackage{amsmath}
\usepackage{amssymb}
\usepackage{mathtools}
\usepackage{amsthm}

\usepackage[capitalize,noabbrev]{cleveref}

\usepackage{hyperref}
\usepackage{url}
\usepackage{enumitem}
\usepackage{booktabs}
\usepackage{comment}
\usepackage{multirow}
\usepackage[font=small,labelfont=bf,tableposition=top]{caption}
\theoremstyle{plain}

\theoremstyle{definition}

\theoremstyle{remark}

\newcommand{\modelname}{\texttt{CHITA}}

\DeclareMathOperator*{\argmin}{arg\,min}
\DeclareMathOperator{\supp}{supp}
\newcommand{\R}{\mathbb{R}}

\usepackage[textsize=tiny]{todonotes}

\icmltitlerunning{}

\begin{document}

\twocolumn[
\icmltitle{Fast as \modelname: Neural Network Pruning with Combinatorial Optimization}



\icmlsetsymbol{equal}{*}

\begin{icmlauthorlist}
\icmlauthor{Riade Benbaki}{MIT}
\icmlauthor{Wenyu Chen}{MIT}
\icmlauthor{Xiang Meng}{MIT}
\icmlauthor{Hussein Hazimeh}{Google Research}
\icmlauthor{Natalia Ponomareva}{Google Research}
\icmlauthor{Zhe Zhao}{Google Research}
\icmlauthor{Rahul Mazumder}{MIT}
\end{icmlauthorlist}

\icmlaffiliation{MIT}{MIT}
\icmlaffiliation{Google Research}{Google Research}

\icmlcorrespondingauthor{Riade Benbaki}{rbenbaki@mit.edu}
\icmlcorrespondingauthor{Wenyu Chen}{wenyu@mit.edu}
\icmlcorrespondingauthor{Xiang Meng}{mengx@mit.edu}
\icmlcorrespondingauthor{Hussein Hazimeh}{hazimeh@google.com}
\icmlcorrespondingauthor{Natalia Ponomareva}{nponomareva@google.com}
\icmlcorrespondingauthor{Zhe Zhao}{zhezhao@google.com}
\icmlcorrespondingauthor{Rahul Mazumder}{rahulmaz@mit.edu}


\icmlkeywords{Machine Learning, ICML}

\vskip 0.3in
]



\printAffiliationsAndNotice{}  

\begin{abstract}
The sheer size of modern neural networks makes model serving a serious computational challenge. A popular class of compression techniques overcomes this challenge by pruning or sparsifying the weights of pretrained networks. While useful, these techniques often face serious tradeoffs between computational requirements and compression quality. 
In this work, we propose a novel optimization-based pruning framework that considers the combined effect of pruning (and updating) multiple weights subject to a sparsity constraint. Our approach, \modelname, extends the classical Optimal Brain Surgeon framework and results in significant improvements in speed, memory, and performance over existing optimization-based approaches for network pruning. \modelname’s main workhorse performs combinatorial optimization updates on a memory-friendly representation of local quadratic approximation(s) of the loss function.  On a standard benchmark of pretrained models and datasets, \modelname~leads to significantly better sparsity-accuracy tradeoffs than competing methods. For example, for MLPNet with only 2\% of the weights retained, our approach improves the accuracy by 63\% relative to the state of the art. Furthermore, when used in conjunction with fine-tuning SGD steps, our method achieves significant accuracy gains over the state-of-the-art approaches.
\end{abstract}

\section{Introduction}


Modern neural networks tend to use a large number of parameters~\citep{devlin2018bert,he2016deep},
which leads to high computational costs during inference. 
A widely used approach to mitigate inference costs is to prune or sparsify pre-trained networks by removing parameters~\citep{blalock2020state}. The goal is to obtain a network with significantly fewer parameters and minimal loss in performance. This makes model storage and deployment cheaper and easier, especially in resource-constrained environments.



Generally speaking, there are two main approaches for neural net pruning: (i) magnitude-based and (ii) impact-based. Magnitude-based heuristic methods~\citep[e.g.,][]{hanson1988comparing,mozer1989using,gordon2020compressing} use the absolute value of weight to determine its importance and whether or not it should be pruned. 
Since magnitude alone may not be a perfect proxy for weight relevance, alternatives have been proposed.
To this end, impact-based pruning methods~\citep[e.g.][]{lecun1989optimal,hassibi1992second,singh2020woodfisher} remove weights based on how much their removal would impact the loss function, often using second-order information on the loss function. Both of these approaches, however, may fall short of  considering the {\emph{joint effect}} of removing (and updating) multiple weights simultaneously. 
The recent method CBS (Combinatorial Brain Surgeon)~\citep{yu2022combinatorial}
is an optimization-based approach that considers the joint effect of multiple weights.
The authors show that CBS leads to a boost in the performance of the pruned models. 
However, CBS can be computationally expensive: it makes use of a local model based on the second-order (Hessian) information of the loss function, which can be prohibitively expensive  in terms of runtime and/or memory (e.g., CBS takes hours to prune a network with 4.2 million parameters).

In this work, we propose \modelname~(\underline{C}ombinatorial \underline{H}essian-free \underline{I}terative \underline{T}hresholding \underline{A}lgorithm), 
an efficient optimization-based 
framework for network pruning at scale. Our approach follows earlier works that consider a local quadratic approximation of the loss function based on the second-order Hessian information. Different from previous works, we make use of a simple yet important observation with which we can avoid computing and storing the Hessian matrix to solve the optimization problem (hence the name ``Hessian-free" in \modelname)---this allows us to address large networks efficiently. Specifically, we propose an equivalent reformulation of the problem as an $\ell_0$-constrained sparse linear regression problem with a data matrix $A\in\R^{n\times p}$, where $p$ is the number of trainable parameters in the original model and  $n\lesssim10^3$ (usually, $p \gg n$) is the number of the sub-samples used in approximating the Hessian. Compared to state-of-the-art approaches that 
consider a dense $p\times p$ matrix  approximation of the Hessian,
our approach leads to a significant reduction in memory usage (up to $10^3$ times for $p=10^6$) without any approximation.

Furthermore, we propose a novel approach to minimize our $\ell_0$ regression reformulation, leveraging active set strategies, better stepsize selection, 
and various methods to accelerate convergence on the selected support. Our proposed approach leads to significant improvements over Iterative Hard Thresholding methods~ \citep[IHT,][]{blumensath2009iterative} commonly used in sparse learning literature. For instance, our framework can prune a network with 4.2M parameters to 80\% sparsity (i.e., 0.84M nonzero parameters) in less than a minute and using less than 20GB of RAM
\footnote{For reference, CBS and Woodfisher would run out of memory in similar circumstances.}.

Since the local quadratic model approximates the loss function only in a small neighborhood of the current solution~\cite{singh2020woodfisher}, we also propose a multi-stage algorithm that updates the local quadratic model during pruning (but without retraining) and solves a more constrained problem in each stage, going from dense weights to sparse ones. Our experiments show that the resulting pruned models have a notably better accuracy compared to that of our single-stage algorithm and other pruning approaches. Furthermore, when used in the gradual pruning setting~\citep{Gale-state-sparsity,singh2020woodfisher,blalock2020state} where re-training between pruning steps is performed, our pruning framework results in significant performance gains compared to state-of-the-art unstructured pruning methods. 

\paragraph{Contributions} Our contributions can be summarized as follows:
\begin{itemize}[leftmargin=*,nosep]
    \item We propose \modelname, an optimization framework  for network pruning based on local quadratic approximation(s) of the loss function. We propose an $\ell_0$-constrained sparse regression reformulation that avoids the pitfalls of storing a large dense Hessian, resulting in a significant reduction in memory usage (we work with an $n\times p$ matrix instead of a $p\times p$ one, with often $n\ll p$).
    \item A key workhorse of \modelname~is a novel IHT-based algorithm to obtain good solutions to the sparse regression formulation. Exploiting problem structure, we propose methods to accelerate convergence and improve pruning performance, such as a new and efficient stepsize selection scheme and rapidly updating weights on the support. This leads to up to 1000x runtime improvement compared to existing network pruning algorithms. 
    \item We show performance improvements across various models and datasets. In particular, 
     \modelname~results in a 98\% sparse (i.e., 98\% of weights in dense model are set to zero) MLPNet with 90\% test accuracy (3\% reduction in test accuracy compared to the dense model), which is a significant improvement over the previously reported best accuracy (55\%) by CBS. As an application of our framework, we use it for gradual pruning and observe notable performance gains against state-of-the-art gradual pruning  approaches.
\end{itemize}

\section{Problem Setup and Related Work}\label{sec:prob-setup-formulation}

In this section we present the general setup with connections to related work---this lays the foundation for our proposed methods discussed in Section~\ref{sec:algo-framework}. 

\subsection{Problem Setup} \label{subsec:setup}
Consider a neural network with an empirical loss function  $
    \mathcal{L}(w)=\frac1N\sum_{i=1}^N\ell_i(w),
$
where $w\in\R^p$ is the vector of trainable parameters, $N$ is the number of data points (samples), and $\ell_i(w)$ is a twice-differentiable function on the $i$-th sample. Given a pre-trained weight vector $ \bar w\in\R^p$, our goal is to set some parameters to zero and possibly update other weights while maintaining the original model's performance (e.g., accuracy) as much as possible. In mathematical terms, given a pre-trained weight $\bar w$ and a target level of sparsity $\tau\in(0,1)$, we aim to construct a new weight vector $w\in\R^p$ that satisfies :
\begin{itemize}[leftmargin=*,nosep]
  \item The loss function at $w$ is as close as possible to the loss before pruning: $\mathcal{L}(w)\approx \mathcal{L}(\bar w)$.
  \item The number of nonzero weights at $w$ respects the sparsity budget\footnote{Here $\ell_0$ norm $\|w\|_0$ denotes the number of nonzero in the vector $w$.}: $\|w\|_0\leq (1-\tau)p$.
\end{itemize}
Similar to \citet{lecun1989optimal,hassibi1992second,singh2020woodfisher}, we use a local quadratic approximation of $\mathcal{L}$ around the pre-trained weight $\bar w$:
\begin{multline}
\label{eqn:local-quadratic}
    \mathcal{L}(w)=\mathcal{L}(\bar w)+\nabla \mathcal{L}(\bar w)^\top (w-\bar w)+ \\ \frac12(w-\bar w)^\top \nabla^2\mathcal{L}(\bar w)(w-\bar w)+O(\|w-\bar w\|^3).
\end{multline}

With certain choices of gradient and Hessian approximations $g\approx\nabla\mathcal{L}(\bar w), H\approx\nabla^2\mathcal{L}(\bar w)$, and ignoring higher-order terms, the loss $\mathcal{L}$ can be locally approximated by:
\begin{equation}\label{eqn:local-quadratic2}
    Q_0(w) :=\mathcal{L}(\bar w)+ g^\top(w-\bar w)+\frac12(w-\bar w)^\top H(w-\bar w).
\end{equation}
Pruning the local approximation $Q_0(w)$ of the network
can be naturally formulated as an optimization problem to minimize 
$Q_0(w)$ subject to a cardinality constraint, i.e., 
\begin{equation}\label{eqn:miqp}
    \min_w ~~ Q_0(w)
    \qquad \text{s.t. }~~~\|w\|_0\le k.
\end{equation}
For large networks, solving Problem \eqref{eqn:miqp} directly (e.g., using iterative optimization methods) is computationally challenging due to the sheer size of the $p \times p$ matrix $H$. In Section \ref{sec:l0_reformulation}, we present an exact, hessian-free reformulation of Problem \eqref{eqn:miqp}, which is key to our scalable approach.

\subsection{Related Work} Impact-based pruning dates back to the work of \citet{lecun1989optimal} where the OBD (Optimal Brain Damage) framework is proposed. This approach, along with subsequent ones \citep{hassibi1992second,singh2020woodfisher,yu2022combinatorial}
make use of local approximation~\eqref{eqn:local-quadratic2}.
It is usually assumed (but not in our work) that $\bar w$ is a local optimum of the loss function, and therefore $g=0$ and $\mathcal{L}(w) \approx \mathcal{L}(\bar w) + \frac12(w-\bar w)^\top H(w-\bar w)$. 
Using this approximation, OBD (Optimal Brain Damage, \citet{lecun1989optimal}) searches for a single weight $i$ to prune with minimal increase of the loss function, while also assuming a diagonal Hessian $H$. If the $i$-th weigth is pruned ($w_i = 0, w_j = \bar w_j \forall j \neq i$), then the loss function increases by $\delta \mathcal{L}_i = \frac{\bar w^2_i}{2 \nabla^2\mathcal{L}(\bar w)}_{ii}$. This represents a score for each weight, and is used to prune weights in decreasing order of their score.
OBS (Optimal Brain Surgeon, \citet{hassibi1992second}) extends this by no longer assuming a diagonal Hessian, and using the optimality conditions to update the un-pruned weights. The authors also propose using the empirical Fisher information matrix, as an efficient approximation to the Hessian matrix. Layerwise OBS \cite{dong2017learning}
proposes to overcome the computational challenge of computing the full (inverse) Hessian needed in OBS by pruning each layer independently, while \citet{singh2020woodfisher} use block-diagonal approximations on the Hessian matrix, which they approximate by the empirical Fisher information matrix on a small subset of the training data ($n \ll N$):
\begin{equation}\label{eqn:Hessian}
    \nabla^2\mathcal{L}(\bar w) \approx H=\frac1n\sum_{i=1}^n\nabla \ell_i(\bar w)\nabla \ell_i(\bar w)^\top.
\end{equation}
While these approaches explore different ways to make the Hessian computationally tractable, they all rely on the OBD/OBS framework of pruning a single weight, and do not to consider the possible interactions that can arise when pruning multiple weights. To this end, \citet{yu2022combinatorial} propose CBS (Combinatorial Brain Surgeon) an algorithm to approximately solve~\eqref{eqn:miqp}.
While CBS shows impressive improvements in the accuracy of the pruned model over prior work, it operates with the full dense $p\times p$ 
Hessian $H$. This limits scalability both in compute time and memory utilization, as $p$ is often in the order of millions and more. 

\paragraph{Choices of gradient approximation $g$.} As mentioned earlier, most previous work assumes that the pre-trained weights $\bar w$ is a local optimum of the loss function $\mathcal{L}$, and thus take the gradient $g=0$.  However, the gradient of the loss function of a pre-trained neural network may not be zero in practice due to early stopping (or approximate optimization)~\cite{yao2007early}. Thus, the WoodTaylor approach~\citep{singh2020woodfisher} proposes to approximate the gradient by the stochastic gradient, using the same samples for estimating the Hessian. Namely,
\begin{equation}\label{eqn:grad-approx}
    g=\frac1n\sum_{i=1}^n\nabla \ell_i(\bar w).
\end{equation}

\paragraph{One-shot and gradual pruning.} 
Generally speaking, \textit{one-shot pruning} methods ~\cite{lecun1989optimal,singh2020woodfisher,yu2022combinatorial} can be followed by a few fine-tuning and re-training steps to recover some of the accuracy lost when pruning. Furthermore, recent work has shown that pruning and re-training in a gradual fashion (hence the name, \textit{gradual pruning}) can lead to big accuracy gains ~\cite{han2015learning,Gale-state-sparsity,ZhuG18}. The work of \citet{singh2020woodfisher} further shows that gradual pruning, when used with well-performing one-shot pruning algorithms, can outperform state-of-the-art unstructured pruning methods. In this paper, we focus on the one-shot pruning problem and show how our pruning framework outperforms other one-shot pruning methods (see Section~\ref{subsec:one-shot-pruning}). We then show that, if applied in the gradual pruning setting, our pruning algorithm outperforms existing approaches (see Section~\ref{subsec:gradual}), establishing new state-of-the-art on MobileNetV1 and ResNet50.

\section{Our Proposed Framework: \modelname}\label{sec:algo-framework}
In this section, we present our algorithmic framework \modelname~(\underline{C}ombinatorial \underline{H}essian-free \underline{I}terative \underline{T}hresholding \underline{A}lgorithm) for pruning a network to specific sparsity level. We formulate sparse network pruning by considering 
$\ell_0$-regression problem(s) and propose scalable algorithms. For example, we can address networks with size $p \approx 10^6, n\approx 10^3, k \approx 10^5$ in less than one minute and using less than 20GB of memory. 
Our basic single-stage algorithm 
is an improved and scalable variant of 
IHT to solve~\eqref{eqn:miqp}.
In Section~\ref{sect:algo-multi}, we propose a multi-stage framework that repeatedly refines the local quadratic approximation and optimizes it (under sparsity constraints) resulting in further performance boosts as shown in Section~\ref{sec:expts}.

\subsection{An $\ell_0$-regression formulation} \label{sec:l0_reformulation}
Our formulation is based on a critical observation that the Hessian approximation in~\eqref{eqn:Hessian} has a low-rank structure: \begin{equation} \label{eq:H_low_rank}
     H=\frac1n\sum_{i=1}^n\nabla \ell_i(\bar w)\nabla \ell_i(\bar w)^\top=\frac1n A^\top A\in\R^{p\times p},
 \end{equation}
 where $A=[\nabla \ell_1(\bar w),\ldots, \nabla \ell_n(\bar w)]^\top\in\R^{n\times p}$ has rank at most $n$ with 
 $10^3 \geq n \ll p$. 
 
 Using observation~\eqref{eq:H_low_rank} and the  gradient expression~\eqref{eqn:grad-approx}, we note that problem \eqref{eqn:miqp} can be equivalently written in the following Hessian-free form:
\begin{equation} \label{eq:q_0_hessian_free}
    \min_w ~~ \frac12\|b- Aw\|^2
    \qquad \text{s.t. }~~~~~\|w\|_0\le k,
 \end{equation}
where $b := A\bar w-e\in\R^n$ and  $e$ is a vector of ones. Furthermore, to improve solution quality (see discussion below), we include a ridge-like regularizer of the form $\|w-\bar w\|^2$ to the objective in~\eqref{eq:q_0_hessian_free}. This leads to the following problem: 
\begin{equation}\label{eqn:main-problem}
    \min_w~Q(w):=\frac12\|b- Aw\|^2+\frac{n\lambda}{2}\|w-\bar w\|^2~~\mathrm{s.t.}~~\|w\|_0\leq k,
\end{equation}
where $\lambda\geq 0$ is a parameter controlling the strength of the ridge regularization. 
Note that our algorithm actually applies to the general form~\eqref{eqn:main-problem}.
Importantly, the regression formulation~\eqref{eqn:main-problem} does not require us to compute or store the full Hessian matrix $H \in \R^{p\times p}$. As discussed in Section~\ref{sect:algo-single}, we only need to operate on the low-rank matrix $A$ throughout our algorithm---this results in substantial gains in terms of both memory consumption and runtime.

\paragraph{Ridge term and choices of $\lambda$.} We observe empirically that the success of our final pruned model depends heavily on the accuracy of the quadratic approximation of the loss function. Since this approximation is local, it is essential to ensure that the weights $w$ during the pruning process are sufficiently close to the initial weights $\bar{w}$. One way\footnote{Another way is to introduce a multi-stage procedure, as explained in Section \ref{sect:algo-multi}.} to achieve this is by including a squared $\ell_2$ penalty, also known as the ridge, on the difference $w-\bar w$.  
This regularization technique does not appear to be explicitly\footnote{It appears to be used implicitly though to obtain an invertible Fisher matrix which is achieved by adding a small diagonal $\lambda_0 I$ to the Fisher matrix.} considered in previous works~\cite{hassibi1992second,singh2020woodfisher,yu2022combinatorial} on pruning using local quadratic approximations.
The usefulness of the regularization term $\lambda$ is further explored in Appendix~\ref{subsubsec:ridge}. We observe that a well-chosen ridge term can help improve the test accuracy on MLPNet by 3\%.

\paragraph{Relation to prior work on $\ell_0$-regression problems.}\label{sect:algo-single} 
There is a substantial literature on algorithms for solving $\ell_0$-regularized linear regression problems.
We provide a brief overview of related work, but it is worth noting that the context of network pruning and the problem-scale we consider here makes our work different from earlier works in $\ell_0$-sparse linear regression. 
\citet{blumensath2009iterative} developed an iterative hard thresholding method, which involves projecting the weights onto the feasible set after each gradient step. \citet{bertsimas2020sparse,hazimeh2022sparse} proposed algorithms to solve sparse regression problems to optimality via branch-and-bound. 
\citet{beck2013sparsity} explore coordinate descent-type algorithms that update one/two coordinates at a time. 
\citet{hazimeh2020fast} propose efficient algorithms based on coordinate descent and local combinatorial optimization that applies to the unconstrained $\ell_0\ell_2$-penalized regression problem. We refer the reader to \citet{hazimeh2022sparse} for a more comprehensive discussion of related work.

In summary, earlier methods for $\ell_0$-regularized linear regression are quite effective at discovering high-quality solutions to Problem~\eqref{eq:q_0_hessian_free} for small to medium-sized problems and require $k$ to be sufficiently small for efficiency. 
However, these methods are not well-suited for tackling large network pruning problems (e.g, $p\sim 10^6$ and $k\sim10^5$) due to slow convergence or expensive per-iteration cost. To address this issue, we propose novel approaches for scalability in Section \ref{sect:algo-single}. Additionally,  we emphasize that \eqref{eqn:main-problem} arises from a local approximation of the true loss $\mathcal L$ around $\bar w$.  Therefore, achieving a high-quality solution for \eqref{eqn:main-problem} alone does not guarantee a pruned network with high accuracy. To this end, we study a multi-stage extension (see Section \ref{sect:algo-multi}) that requires solving several problems of the form~\eqref{eqn:main-problem}.

\subsection{Our proposed algorithm for problem \eqref{eqn:main-problem}}\label{sect:algo-single}
We present the core ideas of our proposed algorithm for Problem~\eqref{eqn:main-problem}, and discuss additional details in Appendix~\ref{app:algo-detail}.

Our optimization framework relies on the IHT algorithm~\citep{blumensath2009iterative,bertsimas2016best} that optimizes~\eqref{eqn:main-problem} by simultaneously updating the support and the weights.
By leveraging the low-rank structure, we can avoid the computational burden of computing the full Hessian matrix, thus reducing complexity.

The basic version of the IHT algorithm can be slow for problems with a million parameters. To improve the computational performance of our algorithm we propose a new line search scheme. Additionally, we use use an active set strategy and schemes to update the weights on the nonzero weights upon support stabilization. Taken together, we obtain notable improvements in computational efficiency and solution quality over traditional IHT, making it a viable option for network pruning problems at scale.

\subsubsection{Structure-aware IHT update}
The IHT algorithm operates by taking a gradient step of size $\tau$ from the current iteration, then projects it onto the set of points with a fixed number of non-zero coordinates through hard thresholding. Specifically, for any vector $x$, let $\mathcal{I}_k(x)$ denote the indices of $k$ components of $x$ that have the largest absolute value. The hard thresholding operator $P_k(x)$ is defined as
$y_{i}=x_{i}$ if $i \in \mathcal{I}_k(x)$, and 
$y_{i}=0$ if $i \notin \mathcal{I}_k(x)$; where $y_i$ is the $i$-th coordinate of $P_{k}(x)$.
IHT applied to problem \eqref{eqn:main-problem} leads to the following update:
\begin{align}\label{eq:iht}
     w^{t+1} &= \textsf{HT}(w^t,k,\tau^s) := P_k\left(w^t-\tau^s \nabla Q(w^t)\right)      \\ 
     &=P_k\left(w^t-\tau^s (A^\top(Ab-w^t)+n\lambda(w^t-\bar w))\right), \nonumber
\end{align}
where $\tau^s>0$ is a suitable stepsize. The computation of $\textsf{HT}(w^t,k,\tau^s)$ involves only matrix-vector multiplications with $A$ (or $A^\top$) and a vector, which has a total computation cost of $O(np)$. This is a significant reduction compared to the $O(p^2)$ cost while using the full Hessian matrix as \citet{singh2020woodfisher,yu2022combinatorial} do.  

\noindent {\bf Active set strategy.} In an effort to further facilitate the efficiency of the IHT method, we propose using an active set strategy, which has been proven successful in various contexts such as~\cite{nocedal1999numerical,friedman2010regularization,hazimeh2020fast}. 
This strategy works by restricting the IHT updates to an \textsl{active set} (a relatively small subset of variables) and occasionally augmenting the active set with variables that violate certain optimality conditions. 
By implementing this strategy, the iteration complexity of the algorithm can be reduced to $O(nk)$ in practice, resulting in an improvement, when $k$ is smaller than $p$. The algorithm details can be found in Appendix~\ref{subsec:active-set}.

\subsubsection{Determining a good stepsize} Choosing an appropriate stepsize $\tau^s$ is crucial for fast convergence of the IHT algorithm. To ensure convergence to a stationary solution, a common choice is to use a constant 
stepsize of $\tau^s=1/L$~\cite{bertsimas2016best,hazimeh2020fast}, where $L$ is the Lipschitz constant of the gradient of the objective function. This approach, while reliable, can lead to conservative updates and slow convergence---refer to Appendix~\ref{subsec:IHT-aggresive} for details. An alternative method for determining the stepsize is to use a backtracking line search, as proposed in~\citet{beck2009fast}. The method involves starting with a relatively large estimate of the stepsize and iteratively shrinking the step size until a sufficient decrease of the objective function is observed. However, this approach requires multiple evaluations of the objective function, which can be computationally expensive. 

\noindent {\bf Our novel scheme.} We propose a novel line search method for determining the stepsize to improve the convergence speed of IHT. Specifically, we develop a method that (approximately) finds the stepsize that leads to the maximum decrease in the objective, i.e., we attempt to solve
\begin{equation} \label{eq:line_search_problem}
\min_{\tau^s\ge 0} ~~ g(\tau^s):=Q\left(P_k\left(w^t-\tau^s \nabla Q(w^t)\right)\right).
\end{equation}
For general objective functions, solving the line search problem (as in \eqref{eq:line_search_problem}) is challenging. However, in our problem, we observe and exploit an important structure: $g(\tau^s)$ is a piecewise quadratic function with respect to $\tau^s$. 
Thus, the optimal stepsize on each piece can be computed exactly, avoiding redundant computations (associated with finding a good stepsize) and resulting in more aggressive updates. In Appendix \ref{subsec:IHT-aggresive}, we present an algorithm that finds a good stepsize by exploiting this structure. Compared to standard line search, our algorithm is more efficient, as it requires fewer evaluations of the objective function and yields a stepsize that results in a steeper descent.


\subsubsection{Additional techniques for scalability}

While the IHT algorithm can be quite effective in identifying the appropriate support, its progress slows down considerably once the support is identified~\cite{blumensath2012accelerated}, resulting in slow convergence. 
We propose two techniques that refine the non-zero coefficients by solving sub-problems to speedup the overall optimization algorithm: (i) Coordinate Descent \citep[CD,][]{bertsekas1997nonlinear,nesterov2012efficiency} that updates each nonzero coordinate (with others fixed) as per a cyclic rule; (ii) Back solve based on Woodbury formula~\cite{max1950inverting} that calculates the optimal solution exactly on a restricted set of size $k$.  We found both (i), (ii) to be important for improving the accuracy of the pruned network. Further details on the strategies (i), (ii) are in Appendix~\ref{subsec:CD}~and~\ref{subsec:backsolve}.


Our single-stage algorithm \modelname~glues together the different pieces discussed above into a coherent algorithm. It takes as input a low-rank matrix $A$, the initial weight $\bar w$ and the $\ell_0$-constraint $k$; and returns a pruned weight $w$ that serves as a good solution to~\eqref{eqn:main-problem}.


\subsection{A multi-stage procedure}\label{sect:algo-multi}

Our single-stage methods (cf Section \ref{sect:algo-single}) lead to high-quality solutions for problem~\eqref{eqn:main-problem}. Compared to existing methods, for a given sparsity level, our algorithms deliver a better objective value for problem \eqref{eqn:main-problem}---for eg, see Figure~\ref{fig:resnet50_obj}. 
However, we note that the final performance (e.g., accuracy) of the pruned network depends heavily on the quality of the local quadratic approximation. This is particularly true when targeting high levels of sparsity (i.e., zeroing out many weights), as the objective function used in \eqref{eqn:main-problem} may not accurately approximate the true loss function $\mathcal L$. 
To this end, we propose a multi-stage procedure named \texttt{CHITA++} that improves the approximation quality by iteratively updating and solving local quadratic models. We use a scheduler to gradually increase the sparsity constraint and take a small step towards higher sparsity in each stage to ensure the validity of the local quadratic approximation. Our multi-stage procedure leverages the efficiency of the single-stage approaches and can lead to pruned networks with improved accuracy by utilizing more accurate approximations of the true loss function. 
For example, our experiments show that the multi-stage procedure can prune ResNet20 to 90\% sparsity in just a few minutes and increases test accuracy from 15\% to 79\% compared to the single-stage method.
Algorithm \ref{alg:multi} presents more details on \texttt{CHITA++}.

Our proposed multi-stage method differs significantly from the gradual pruning approach described in \citet{han2015learning}. While both methods involve pruning steps, the gradual pruning approach also includes fine-tuning steps in which SGD is applied to further optimize the parameters for better results. However, these fine-tuning steps can be computationally expensive, usually taking days to run. In contrast, our proposed multi-stage method is a one-shot pruning method and only requires constructing and solving Problem \eqref{eqn:main-problem} several times, resulting in an efficient and accurate solution. This solution can then be further fine-tuned using SGD or plugged into the gradual pruning framework, something we explore in Section \ref{subsec:gradual}.

\begin{algorithm}[h]
\begin{algorithmic}[1]
\REQUIRE Pre-trained weights $\bar w$, a target sparsity level $\tau$, number of stages $f$.
\STATE \textbf{Initialization:} Construct a increasing sequence of sparsity parameters $\tau_1,\tau_2,\dots,\tau_f=\tau$; and set $w^0=\bar w$ 
\FOR {$t=1,2,\ldots,f$}
\STATE At current solution $w^{t-1}$, calculate the gradient based on a batch of $n$ data points and construct the matrix $A$ given in \eqref{eqn:Hessian}.
\STATE Obtain a solution $w^t$ to problem~\eqref{eqn:main-problem} by invoking \texttt{CHITA}$(A, \bar w,k)$ with $\bar w=w^{t-1}$ and number of nonzeros $k=\lfloor (1-\tau_t)p\rfloor$.
\ENDFOR
\end{algorithmic}
\caption{\texttt{CHITA++}: a multi-stage pruning procedure}
\label{alg:multi}
\end{algorithm}



\section{Experimental Results}\label{sec:expts}
We compare our proposed framework with existing approaches, for both one-shot and gradual pruning.
\subsection{One shot pruning}\label{subsec:one-shot-pruning}
We start by comparing the performance of our methods: \texttt{CHITA} (single-stage) and \texttt{CHITA++} (multi-stage) with several existing state-of-the-art one-shot pruning techniques on various pre-trained networks. We use the same experimental setup as in recent work \cite{yu2022combinatorial,singh2020woodfisher}. The existing pruning methods we consider include MP~\citep[Magnitude Pruning,][]{mozer1989using}, WF~\citep[WoodFisher,][]{singh2020woodfisher}, CBS~\citep[Combinatorial Brain Surgeon,][]{yu2022combinatorial} and M-FAC~\citep[Matrix-Free Approximate Curvature,][]{frantar2021m}. The pre-trained networks we use are MLPNet (30K parameters) trained on MNIST~\cite{lecun1998gradient}, ResNet20~(\citealp{he2016deep}, 200k parameters) trained on CIFAR10~\cite{krizhevsky2009learning}, and MobileNet (4.2M parameters) and ResNet50~(\citealp{he2016deep}, 22M parameters) trained on ImageNet~\cite{deng2009imagenet}. 
For further details on the choice of the Hessian approximation, we refer the reader to 
Appendix~\ref{subsec:stratified-block}. Detailed information on the experimental setup and reproducibility can be found in Appendix~\ref{subsecsec:expt-one-shot}.

\subsubsection{Runtime comparison} 

{\footnotesize
\begin{figure}
     \centering

\includegraphics[width=0.9\columnwidth]{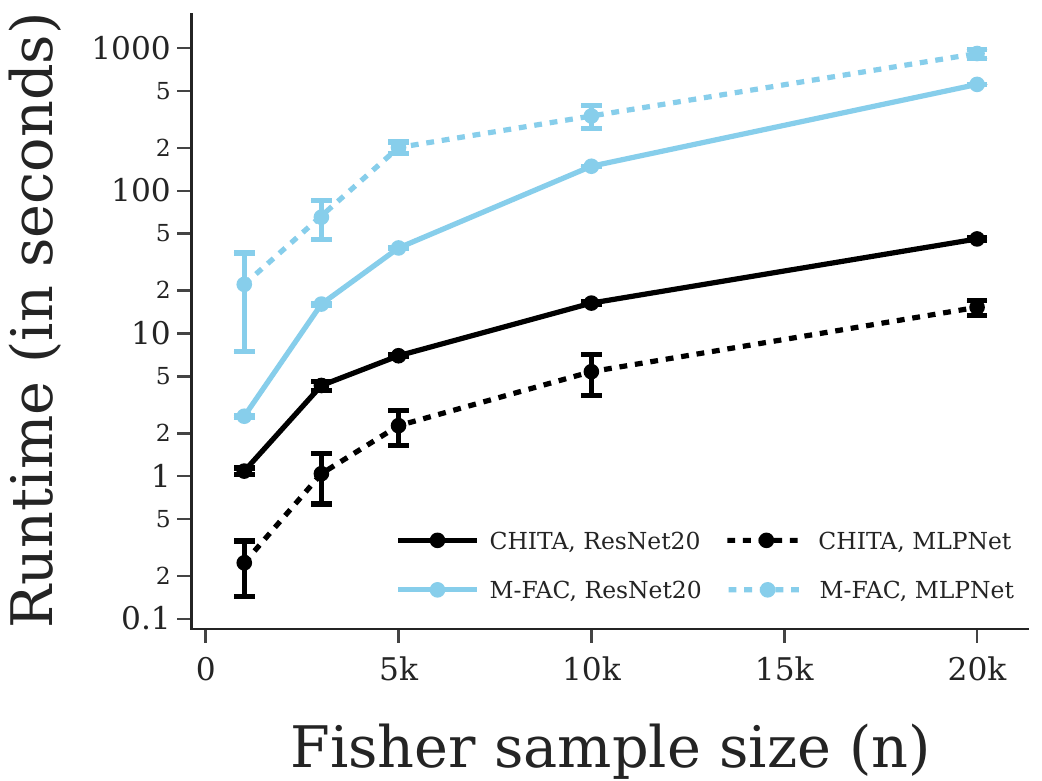}
     \caption{Runtime comparison between our single-stage approaches and M-FAC (the fastest among the competitive methods) while pruning MLPNet and ResNet20 to 90\% sparsity level (90\% of the entries are zero). Note that Woodfisher and CBS are at least 1000 times slower than M-FAC. The error bar represents the standard error over four runs. \texttt{CHITA} here uses IHT to find a support and performs a back-solve on the found support.}
     \label{fig:runtime}
     \vspace{-3mm}
\end{figure}}

Recent works that use the empirical Fisher information matrix for pruning purposes \citep{singh2020woodfisher,yu2022combinatorial} show that using more samples for Hessian and gradient approximation results in better accuracy. Our experiments also support this conclusion.  However, most prior approaches become computationally prohibitive as sample size $n$ increases. As an example, the Woodfisher and CBS algorithms require hours to prune a MobileNet when $n$ is set to 1000, and their processing time increases at least linearly with $n$. In contrast, our method has been designed with efficiency in mind, and we have compared it to M-FAC, a well-optimized version of Woodfisher that is at least 1000 times faster. The results, as depicted in Figure \ref{fig:runtime}, demonstrate a marked improvement in speed for our algorithm, with up to 20 times faster performance.

\subsubsection{Accuracy of the pruned models} 
\begin{table}[ht!]
    \centering
    \resizebox{\columnwidth}{!}
    {\begin{tabular}{c|c|ccc|cc}
    \toprule
\footnotesize Network & Sparsity & MP & WF & CBS & \modelname & \texttt{CHITA++} \\
\midrule
\multirow{7}{*}{
\begin{minipage}{2cm}
\begin{center}
    MLPNet\\
    on MNIST\\
    (93.97\%)
\end{center}
\end{minipage}
} &0.5 & 93.93 & 94.02 & 93.96 & 93.97 (±0.03) & \textbf{95.97} (±0.05) \\
&0.6 & 93.78 & 93.82 & 93.96 & 93.94 (±0.02) & \textbf{95.93} (±0.04) \\
&0.7 & 93.62 & 93.77 & 93.98 & 93.80 (±0.01) & \textbf{95.89} (±0.06) \\
&0.8 & 92.89 & 93.57 & 93.90 & 93.59 (±0.03) & \textbf{95.80} (±0.03) \\
&0.9 & 90.30 & 91.69 & 93.14 & 92.46 (±0.04) & \textbf{95.55} (±0.03) \\
&0.95 & 83.64 & 85.54 & 88.92 & 88.09 (±0.24) & \textbf{94.70} (±0.06) \\
&0.98 & 32.25 & 38.26 & 55.45 & 46.25 (±0.85) & \textbf{90.73} (±0.11) \\

\midrule
\multirow{7}{*}{
\begin{minipage}{2cm}
\begin{center}
    ResNet20\\
    on CIFAR10\\
    (91.36\%)
\end{center}
\end{minipage}
}&0.3 & 90.77 & \textbf{91.37} & 91.35 & \textbf{91.37 (±0.04)} & 91.25 (±0.08) \\
&0.4 & 89.98 & 91.15 & \textbf{91.21} & 91.19 (±0.05) & \textbf{91.20} (±0.05) \\
&0.5 & 88.44 & 90.23 & 90.58 & 90.60 (±0.07) & \textbf{91.04} (±0.09) \\
&0.6 & 85.24 & 87.96 & 88.88 & 89.22 (±0.19) & \textbf{90.78} (±0.12) \\
&0.7 & 78.79 & 81.05 & 81.84 & 84.12 (±0.38) & \textbf{90.38} (±0.10) \\
&0.8 & 54.01 & 62.63 & 51.28 & 57.90 (±1.04) & \textbf{88.72} (±0.17) \\
&0.9 & 11.79 & 11.49 & 13.68 & 15.60 (±1.79) & \textbf{79.32} (±1.19) \\

\midrule
\multirow{6}{*}{
\begin{minipage}{2cm}
\begin{center}
    MobileNetV1\\
    on ImageNet\\
    (71.95\%)
\end{center}
\end{minipage}
}&0.3 & 71.60 & \textbf{71.88} & \textbf{71.88} & \textbf{71.87} (±0.01) & {71.86} (±0.02) \\
&0.4 & 69.16 & 71.15 & 71.45 & 71.50 (±0.02) & \textbf{71.61} (±0.02) \\
&0.5 & 62.61 & 68.91 & 70.21 & 70.42 (±0.02) & \textbf{70.99} (±0.04) \\
&0.6 & 41.94 & 60.90 & 66.37 & 67.30 (±0.03) & \textbf{69.54} (±0.01) \\
&0.7 & 6.78 & 29.36 & 55.11 & 59.40 (±0.09) & \textbf{66.42} (±0.03) \\
&0.8 & 0.11 & 0.24 & 16.38 & 29.78 (±0.18) & \textbf{47.45} (±0.25) \\
\bottomrule
 \end{tabular} }
\vspace{10pt}
\caption{The pruning performance (model accuracy) of various methods on MLPNet, ResNet20, MobileNetV1. As to the performance of MP, WF, and CBS, we adopt the results reported in \citet{yu2022combinatorial}. We take five runs for our single-stage (\texttt{CHITA}) and multi-stage (\texttt{CHITA++}) approaches and report the mean and standard error~(in the brackets). The best accuracy values  (significant) are highlighted in bold. Here sparsity denotes the fraction of zero weights in convolutional and dense layers.}
\label{tab:accuracy}
\end{table}
\noindent {\bf Comparison against state-of-the-art.}
Table~\ref{tab:accuracy} compares the test accuracy of MLPNet, ResNet20 and MobileNetV1 pruned to different sparsity levels. Our single-stage method achieves comparable results to other state-of-the-art approaches with much less time consumption. The multi-stage method 
(\texttt{CHITA++}) outperforms other methods by a large margin, especially with a high sparsity rate. 

\begin{figure}[!h]
    \centering
    \subfigure[Test accuracy for one-shot pruning on ResNet50. ]{
         \centering
\includegraphics[width=0.78\columnwidth]{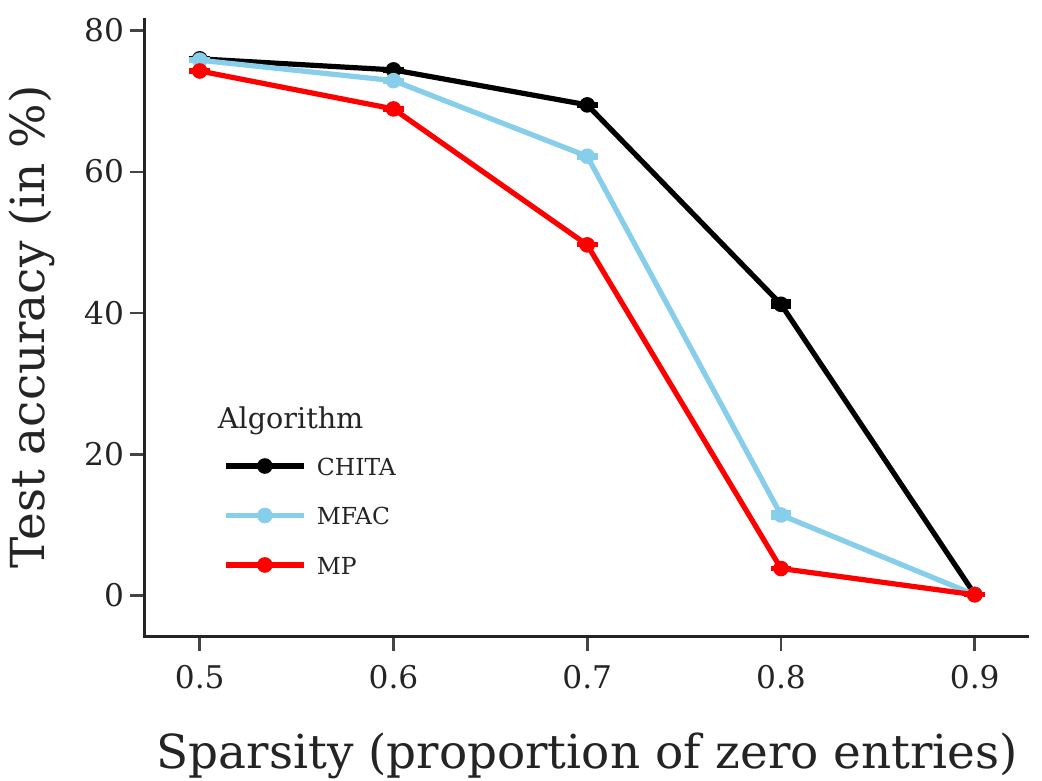}
  \label{fig:resnet50_acc}
     }
    \subfigure[The objective value in \eqref{eqn:main-problem} for pruning ResNet50.]{
         \centering
\includegraphics[width=0.78\columnwidth]{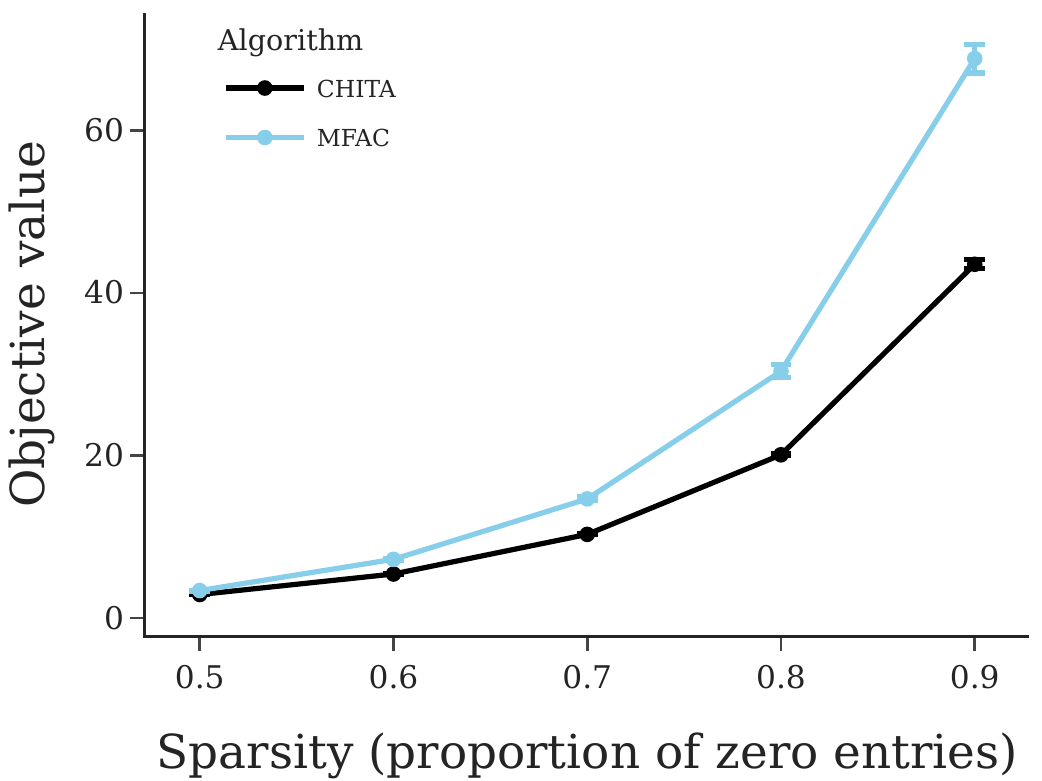}
         \label{fig:resnet50_obj}
     }
    \caption{One-shot pruning on ResNet50 (Dense accuracy is 77.01\%). The error bars are over four runs. For a fair comparison, both \modelname~and M-FAC use the same hyperparameters and the same training samples for Hessian and gradient approximation.}
    \label{fig:resnet50_oneshot}
  \end{figure}

\noindent {\bf One-shot pruning on ResNet50.}
We further compare our approach to competing methods on ResNet50, an even larger network where some pruning algorithms, like CBS, do not scale. In Figure~\ref{fig:resnet50_oneshot}, we evaluate the performance of our algorithm in comparison to M-FAC and Magnitude Pruning (MP) using two metrics: test accuracy and the final objective value of the $\ell_0$-constrained problem~\eqref{eqn:main-problem}. As both M-FAC and our algorithm aim to minimize this objective, it can be used  to judge the efficacy of our model in solving problem \eqref{eqn:main-problem}. As seen in the figure, our approach achieves a lower objective value, and in this case, it also results in a better test accuracy for the final pruned network. 

\begin{figure}[!h]
    \centering
\includegraphics[width=0.8\columnwidth]{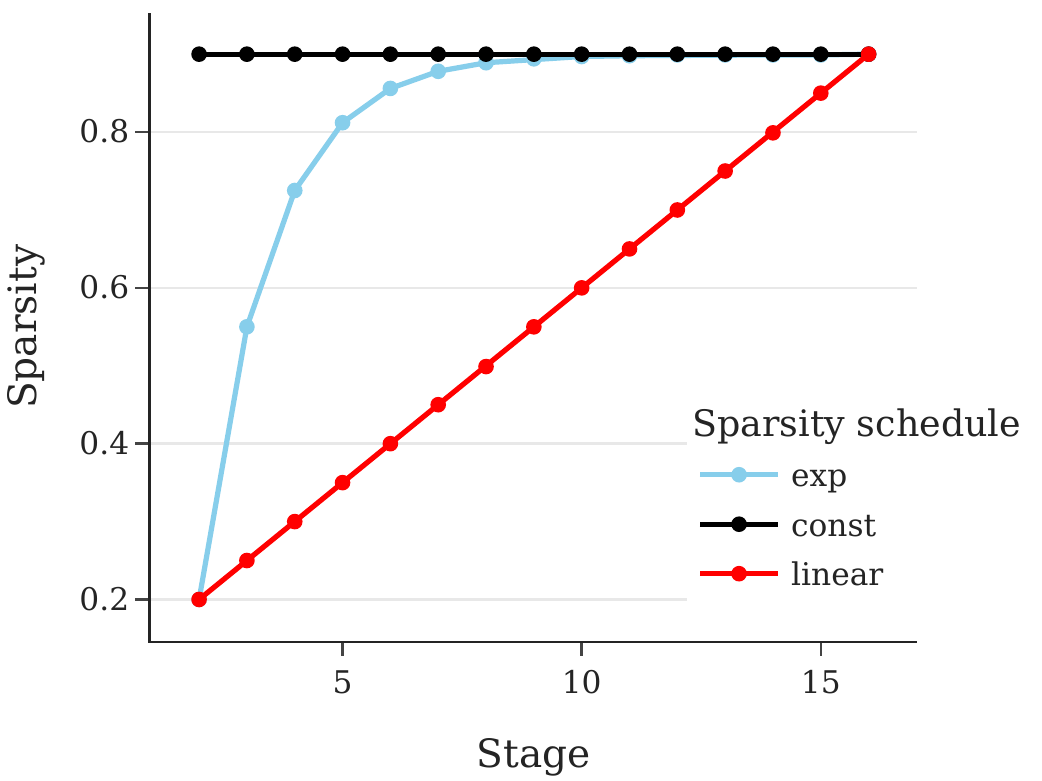}
    \caption{Three different sparsity schedules: exponential, linear, and constant schedules.}
    \label{fig:iter_schedules}
\end{figure}

\noindent {\bf Sparsity schedule in multi-stage procedure.}\label{par:sparsity-schedule}
We study the effect of the sparsity schedule (i.e., choice of $\tau_1\le \tau_2\le \cdots \le \tau_f=\tau$ in Algorithm \ref{alg:multi}) on the performance of \texttt{CHITA++}. We compare test accuracy of three different schedules:~(i)~exponential mesh,~(ii)~linear mesh, and ~(iii)~constant mesh. For these schedules, $f$ is set to be $15$. For the first two meshes, $\tau_1$ and $\tau_{15}$ are fixed as $0.2$ and $0.9$, respectively. As shown in Figure \ref{fig:iter_schedules}, the exponential mesh computes $\tau_2,\dots,\tau_{14}$ by drawing an exponential function, while the linear mesh adopts linear interpolation (with $\tau_1$ and $\tau_{15}$ as endpoints) to determine $\tau_2,\dots,\tau_{14}$ and the constant mesh has $\tau_1 = \tau_2 =\cdots = \tau_{15}$. 

\begin{figure}[!h]
         \centering
\includegraphics[width=\columnwidth]{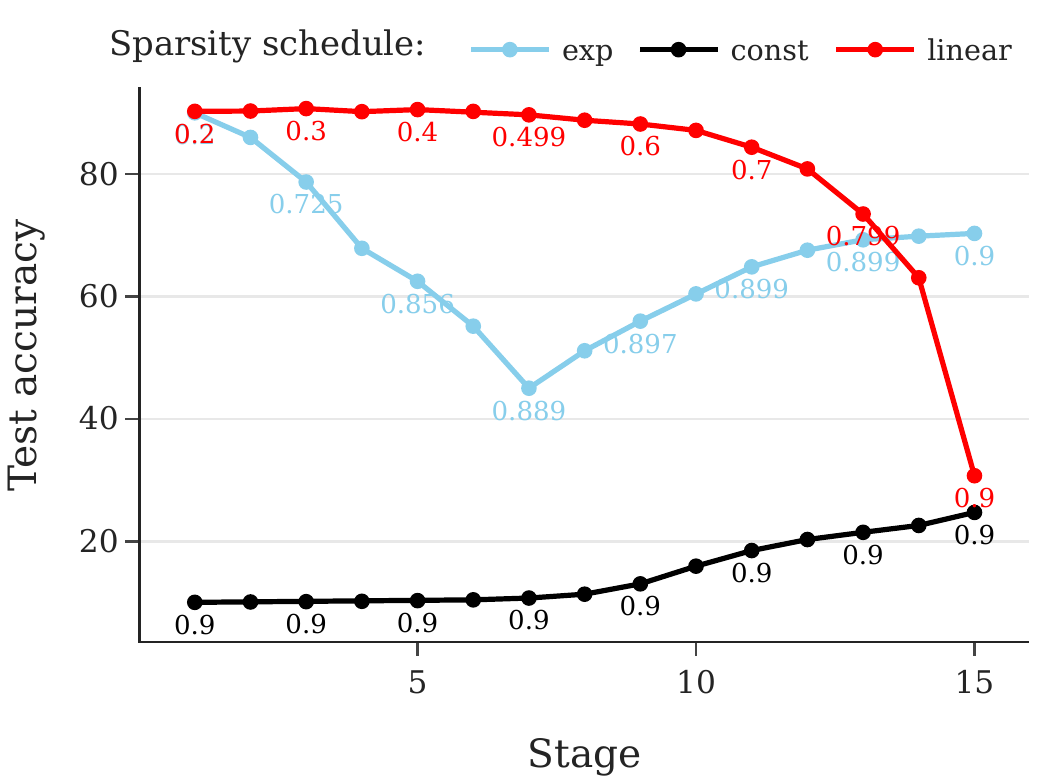}
    
\caption{Comparison of test accuracy using \texttt{CHITA++} with 15 stages, pruning a ResNet20 model to a 90\% sparsity rate, under different sparsity schedules. Text around the point indicates the current sparsity level of the point.}
     \label{fig:training_loss}
     \vspace{-5mm}
  \end{figure}

Figure \ref{fig:training_loss} plots the test accuracy of the three schedules over the number of stages. We observe that the linear mesh outperforms the  exponential mesh in the first few iterations, but its performance drops dramatically in the last two iterations. The reason is that in high sparsity levels, even a slight increase in the sparsity rate leads to a large drop in accuracy. Taking small ``stepsizes'' in high sparsity levels allows the exponential mesh to fine-tune the weights in the last several stages and achieve good performance. 

\paragraph{Additional ablation studies.}
We perform additional ablation studies to further evaluate the performance of our method. These studies mainly focus on the effect of the ridge term (in Appendix \ref{subsubsec:ridge}), and the effect of the first-order term (in Appendix \ref{subsubsec:first-order}).

\subsection{Performance on gradual pruning}\label{subsec:gradual}
To compare our one-shot pruning algorithms against more unstructured pruning methods, we plug \texttt{CHITA} into a gradual pruning procedure \cite{Gale-state-sparsity}, following the approach in \citet{singh2020woodfisher}. Specifically, we alternate between pruning steps where a sparse weight is computed and fine-tuning steps on the current support via Stochastic Gradient Descent (SGD). To obtain consistent results,
we start from the same pre-trained weights used in \citet{Kusupati2020STR}, and
re-train for 100 epochs using SGD during fine-tuning steps, similarly to \citet{Kusupati2020STR,singh2020woodfisher}. We compare our approach against Incremental~\cite{ZhuG18}, STR~\cite{Kusupati2020STR}, Global Magnitude~\cite{singh2020woodfisher}, WoodFisher~\cite{singh2020woodfisher}, GMP~\cite{Gale-state-sparsity}, Variational Dropout ~\cite{10.5555/3305890.3305939}, RIGL~\cite{evci2020rigging}, SNFS~\cite{dettmers2020sparse} and DNW~\cite{DNW2019}. Further details on training procedure can be found in Appendix~\ref{subsubsec:expt-setup-gradual pruning}. 

\begin{table}[h]
\centering
\resizebox{\columnwidth}{!}
{\begin{tabular}{ccccc}
\toprule \footnotesize
       & Sparsity &   Pruned    & Relative Drop (\%)  & Remaining    \\
 Method& (\%) & Accuracy & $\frac{Pruned-Dense}{Dense}$                 & \# of params \\ 
\midrule
Incremental & 74.11 & 67.70 & -4.11 & 1.09 M \\
STR & 75.28 & 68.35 & -5.07 & 1.04 M \\
Global Magnitude & 75.28 & 69.90 & -2.92 & 1.04 M \\
WoodFisher & 75.28 & 70.09 & -2.65 & 1.04 M \\
\textbf{\modelname} & 75.28 & \textbf{71.11} & \textbf{-1.23} &  1.04 M \\
\midrule
Incremental & 89.03 & 61.80 & -12.46 & 0.46 M \\
STR & 89.01 & 62.10 & -13.75 & 0.46 M \\
Global Magnitude & 89.00 & 63.02 & -12.47 & 0.46 M \\
WoodFisher & 89.00 & 63.87 & -11.29 & 0.46 M \\
\textbf{\modelname} & 89.00 & \textbf{67.68} & \textbf{-6.00} &  0.46 M \\
\bottomrule
\end{tabular}}
\caption{\footnotesize Results of gradually pruning MobilenetV1 in 75\% and 89\% sparsity regimes, comparing \modelname\ to other baselines (Dense accuracy: 72.00\%). We also include the relative drop in accuracy to account for different methods starting from different dense weights. \modelname\  numbers are averaged across two runs. Numbers for other baselines are taken from \citet{singh2020woodfisher}.  }
\label{table:gradualpruning}
\vspace{-5mm}
\end{table}

\noindent {\bf MobileNetV1.} We start by pruning MobileNetV1 (4.2M parameters). As Table~\ref{table:gradualpruning} demonstrates, \modelname\ results in significantly more accurate pruned models than previous state-of-the-art approaches at sparsities 75\% and 89\%, with only 6\% accuracy loss compared to 11.29\%, the previous best result when pruning to a sparsity of 89\%. 

\noindent {\bf ResNet50.}
Similarly to MobileNetV1, \modelname \ improves test accuracy at sparsity levels 90\%, 95\%, and 98\% compared to all other baselines, as Table \ref{table:resnet50gradual} shows. This improvement becomes more noticeable as we increase the target sparsity, with \modelname ~producing a pruned model with 69.80\% accuracy compared to 65.66\%, the second-best performance, and previous state-of-the-art.
\begin{table}[h]
\centering
\resizebox{\columnwidth}{!}
{\begin{tabular}{cccccc}
\midrule \footnotesize
                    & Sparsity &   Pruned    & Relative Drop (\%) & Remaining    \\
Method              & (\%) & Accuracy & $\frac{Pruned-Dense}{Dense}$                 & \# of params \\ 
\midrule
GMP + LS       & 90.00 & 73.91 & -3.62 & 2.56 M \\
Variational Dropout & 90.27 & 73.84 & -3.72 & 2.49 M \\
RIGL + ERK          & 90.00 & 73.00 & -4.94 & 2.56 M \\
SNFS + LS           & 90.00 & 72.90 & -5.32 &   2.56 M \\
STR                 & 90.23 & 74.31 & -3.51 &   2.49 M \\
Global Magnitude    & 90.00 & 75.20 & -2.42 &   2.56 M \\
DNW                 & 90.00 & 74.00 & -4.52 &   2.56 M \\
WoodFisher          & 90.00 & 75.21 & -2.34 &  2.56 M \\
\textbf{\modelname} & 90.00 & \textbf{75.29} & \textbf{-2.23} &  2.56 M \\ 
\midrule
GMP & 95.00 & 70.59 & -7.95  & 1.28 M \\
Variational Dropout & 94.92 & 69.41 & -9.49  & 1.30 M \\
Variational Dropout & 94.94 & 71.81 & -6.36  & 1.30 M \\
RIGL + ERK & 95.00 & 70.00 & -8.85  & 1.28 M \\
DNW & 95.00 & 68.30 & -11.31  & 1.28 M \\
STR & 94.80 & 70.97 & -7.84  & 1.33 M \\
STR & 95.03 & 70.40 & -8.58  & 1.27 M \\
Global Magnitude & 95.00 & 71.79 & -6.78  & 1.28 M \\
WoodFisher & 95.00 & 72.12 & -6.35  & 1.28 M \\
\textbf{\modelname} &95.00 & \textbf{73.46} & \textbf{-4.61}  &1.28 M \\
\midrule
GMP + LS            & 98.00 & 57.90 & -24.50  & 0.51 M \\
Variational Dropout & 98.57 & 64.52 & -15.87  & 0.36 M \\
DNW                 & 98.00 & 58.20 & -24.42  & 0.51 M \\
STR                 & 98.05 & 61.46 & -20.19  & 0.50 M \\
STR                 & 97.78 & 62.84 & -18.40  & 0.57 M \\
Global Magnitude    & 98.00 & 64.28 & -16.53  & 0.51 M \\
WoodFisher          & 98.00 & 65.55 & -14.88  & 0.51 M \\
\textbf{\modelname}  & 98.00 & \textbf{69.80} & \textbf{-9.36}   & 0.51M \\
\midrule
\end{tabular}}
\caption{Results of gradually pruning a ResNet50 network in the 90\%, 95\%, and 98\% sparsity regimes, comparing \modelname\ to other state-of-the-art methods (Dense accuracy: 77.01\%). We also include the relative drop in accuracy to account for different methods starting from different dense weights. \modelname\  numbers are averaged across two runs. Numbers for other baselines are taken from \citet{singh2020woodfisher}.}
\label{table:resnet50gradual}
\vspace{-10.9mm}
\end{table}
\section{Conclusion}
In this work we have presented an efficient network pruning framework \modelname~, which is based on a novel, hessian-free $\ell_0$-constrained regression formulation and  combinatorial optimization techniques. Our single-stage methods demonstrate comparable results to existing methods while achieving a significant improvement in runtime and reducing memory usage. Furthermore, by building upon the single-stage methods, our multi-stage approach is capable of achieving even further improvements in model accuracy. Additionally, we have demonstrated that incorporating our pruning methods into existing gradual pruning frameworks results in sparse networks with state-of-the-art accuracy.





\section*{Acknowledgements}

This research is supported in part by grants from the Office of Naval Research (N000142112841 and N000142212665), and Google. 
We thank Shibal Ibrahim for helpful discussions. We also thank Thiago Serra and Yu Xin for sharing with us code from their CBS paper~\cite{yu2022combinatorial}.

\bibliography{references}
\bibliographystyle{icml2022}

\newpage
\appendix
\onecolumn
\section{Algorithmic details}\label{app:algo-detail}

\subsection{IHT with aggressive stepsize}\label{subsec:IHT-aggresive}



\paragraph{Challenges of stepsize choice}
Choosing an appropriate stepsize $\tau^s$ is crucial to achieving a faster convergence rate.  In theory, setting $\tau^s$ as the constant $1/L$ in \eqref{eq:iht} is a common choice in the literature to ensure the convergence to a stationary solution~\cite{bertsimas2016best,hazimeh2020fast}, where $L$ is the Lipschitz constant of the gradient of $Q(w)$. i.e., $\|\nabla Q(\alpha)-\nabla Q(\beta)\|\le L\|\alpha-\beta\|$ for all $\alpha,\,\beta\in \mathbb{R}^p$. Since $Q$ is a quadratic objective, this quantity $L$ is given by $L=n\lambda+\|A\|_{2}^2$, where $\|A\|_{2}$ is the maximum singular value of $A$. This quantity could be substantial when $p$ is large, leading to very conservative updates, sometimes negligible. Moreover, the computation of $L$ itself may involve a few power iterations or a randomized SVD algorithm, which could be as costly as several IHT updates. An alternative method for determining the stepsize is to use a backtracking line search, as proposed in~\citet{beck2009fast}. The method involves starting with a relatively large estimate of the stepsize and iteratively shrinking the step size until a sufficient decrease of the objective function is observed. However, this method requires multiple evaluations of the quadratic objective, which can be computationally expensive.

\begin{figure*}[h]
     \centering
     \includegraphics[width=0.8\columnwidth]{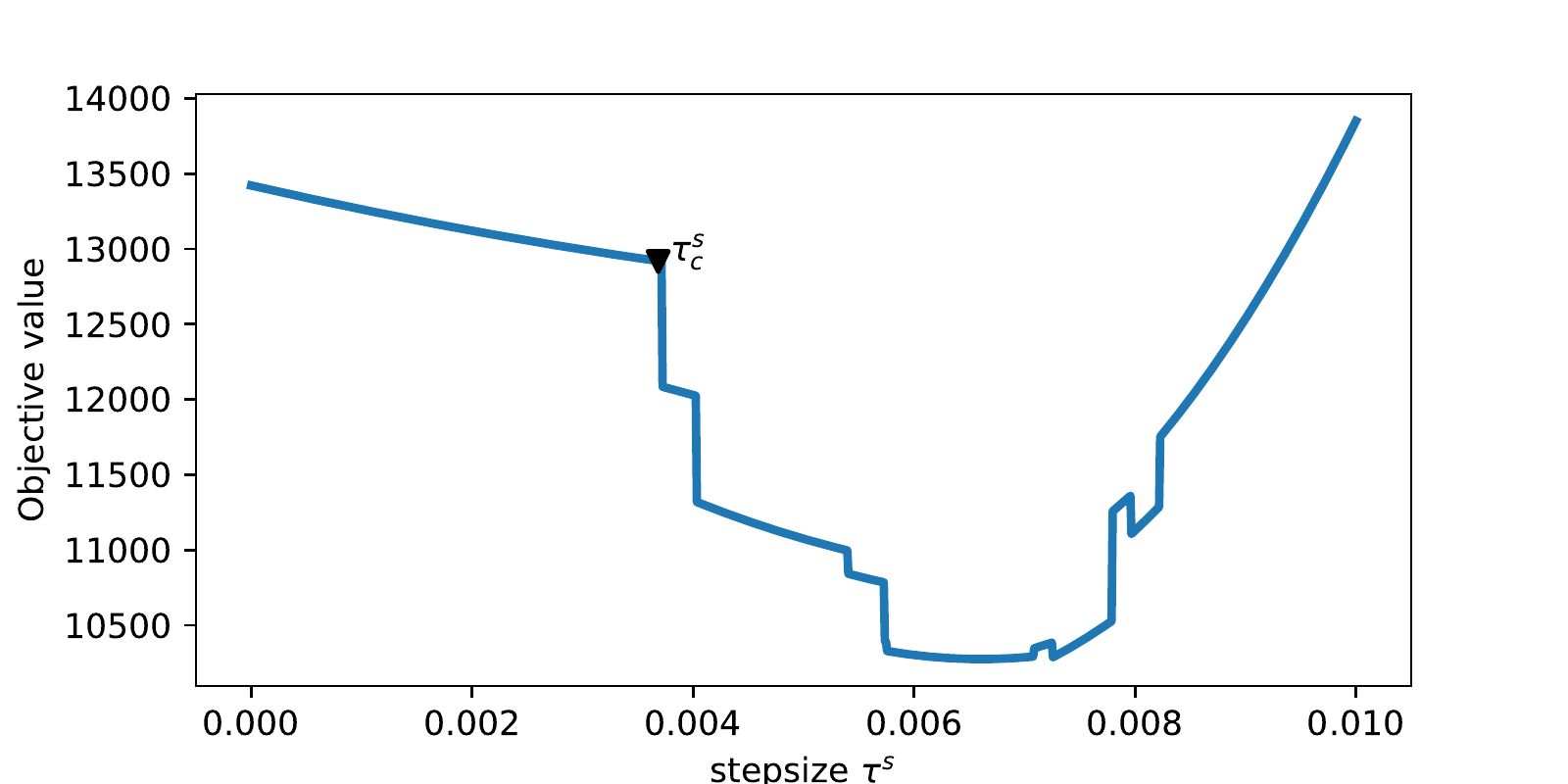}
     \caption{An example of $g(\tau^s)$ for $n=100,p=200$ and $k=10$. The black triangle denotes the first breaking point of $g(\tau^s)$, with its x-coordinate represented by $\tau_c^s$.}
     \label{fig:stepsize}
\end{figure*}

\paragraph{Our novel scheme}
We propose a new line search strategy to efficiently determine an aggressive stepsize to address the issue of slow updates in the IHT algorithm. Note that the problem of finding the best stepsize can be written as the following one-dimensional problem 
\begin{equation}
\min_{\tau^s \ge 0} g(\tau^s):=Q\left(P_k\left(w^t-\tau^s \nabla Q(w^t)\right)\right).
\end{equation}
Since $P_k$ is a piecewise function, $g(\tau^s)$ is a univariate piecewise quadratic function which is generally non-convex, as illustrated in Figure \ref{fig:stepsize}.  Our key observation is that the first breaking point of $g(\tau^s)$ and the optimal stepsize on the first piece can be computed easily. More specifically, denote by $\tau_c^s$ the first breaking point of $g(\tau^s)$. Namely, $\tau_c^s$ is the largest value of $\tau'$ such that the hard thresholding based on $\tau^s\in[0,\tau']$ does not change the support, i.e. $\mathcal{I}_k(w^t)=\mathcal{I}_k(w^t-\tau^s\nabla Q(w^t)),~\forall\tau^s\in[0,\tau']$.  Let us denote $\mathcal{S} := \text{supp}(w)$. In the case where $|\mathcal{S}|=k$, $\tau_c^s$ can be computed in closed form using
\vspace{2mm}
\begin{equation}
    \tau_c^s= \min_{i\in \mathcal{S}} \left\{\frac{|w_i^t|}{\max\{\max_{j\notin \mathcal{S}} |[\nabla Q(w^t)]_j|-[\nabla Q(w^t)]_i\text{sign}(w_i^t[\nabla Q(w^t)]_i),0_+\}}\right\}.\vspace{2mm}
\end{equation}
As previously established, over the interval $\tau^s\in[0,\tau_c^s]$, the function $g(\tau^s)=Q\left(w^t-\tau^s \nabla Q(w^t)\right)$ is a quadratic function. Let us denote by $\tau_m^s$ the optimal value of $\tau^s$ that minimizes $g(\tau^s)$ within the interval $[0,\tau_c^s]$. It is straightforward to see that $\tau_m^s$ can be computed in closed form with the same computational cost as a single evaluation of the quadratic objective function.

If $\tau_m^s<\tau_c^s$, then the optimal value of $\tau^s$ lies within the first quadratic piece. Practically, we have found that in this case $\tau_m^s$ is often also the global minimum of $g(\tau^s)$. Therefore, we can confidently take the stepsize as $\tau^s=\tau_m^s$. Otherwise if $\tau_m^s=\tau_c^s$, then we know that $g(\tau^s)$ is monotonically decreasing on the interval $[0,\tau_c^s]$. This implies that $g(\tau^s)$ would likely continue to decrease as $\tau^s$ becomes larger than $\tau_c^s$. As a result, we perform a line search by incrementally increasing the value of $\tau^s$ by a factor of $\gamma>1$ starting from $\tau_c^s$ to approximate the stepsize that results in the steepest descent. The above procedure is summarized in Algorithm~\ref{alg:stepsize}.



Our proposed scheme offers a significant improvement in efficiency compared to standard backtracking line search by eliminating redundant steps on the quadratic piece of $g(\tau^s)$ over $[0,\tau_c^s]$. Additionally, our method directly computes the optimal stepsize on the first piece, which in many cases, results in a greater reduction in the objective function when compared to the standard backtracking line search.

Finally, we note that during line search, it is always possible to find the piece of the quadratic function to which the current stepsize $\tau^s$ belongs, say $[\tau_l^s,\tau_u^s]$, and calculate the optimal stepsize over that piece with small extra costs to further improve the line search. But we find it unnecessary in practice as the line search procedure usually terminates in a few steps.

\begin{algorithm}[h]
\begin{algorithmic}[1]
\REQUIRE $w^t$, $k$, $\gamma>1$
\STATE Compute $\tau_c^s$ as the first breaking point of $g(\tau^s)$ in closed form. \COMMENT{Time complexity $O(p)$}
\STATE Compute 
\begin{equation}
    \tau_m^s=\argmin_{\tau^s\in [0,\tau_c^s]} g(\tau^s):= Q\left(P_k\left(w^t-\tau^s \nabla Q(w^t)\right)\right)
\end{equation}
as the optimal stepsize on the first piece of $g(\tau^s)$ in closed form.  \COMMENT{Time complexity $O(nk)$}
\IF {$\tau_m^s<\tau_c^s$}
\STATE $\tau^s\gets\tau_m^s$.
\ELSE
\STATE $g_{\text{best}}\gets g(\tau_c^s)$, and $\tau^s\gets\tau_c^s$.
\WHILE {$g_{\text{best}} > g(\gamma\tau^s)$}
\STATE $g_{\text{best}}\gets g(\gamma\tau^s)$, and $\tau^s\gets \gamma\tau^s$.
\ENDWHILE
\ENDIF
\end{algorithmic}
\caption{A novel scheme to determine the stepsize $\tau^s$}
\label{alg:stepsize}
\end{algorithm}

\subsection{Cyclic coordinate descent}\label{subsec:CD}

Although IHT does well in identifying and updating the support, we observe that it makes slow progress in decreasing the objective in experiments. To address this issue, we use cyclic coordinate descent \citep[CD,][]{bertsekas1997nonlinear,nesterov2012efficiency} with full minimization in every nonzero coordinate to refine the solution on the support. CD-type methods are widely used for solving huge-scale optimization problems in statistical learning, especially those problems with sparsity structure, due to their inexpensive iteration updates and capability of exploiting problem structure, such as Lasso~\cite{friedman2010regularization} and $L_0L_2$-penalized regression \cite{hazimeh2020fast}. 

Our cyclic CD updates each nonzero coordinate (with others fixed) as per a cyclic rule, and skips any coordinate with zero value to avoid violating the $\ell_0$ constraint. With a feasible initialization $w^t$ and a coordinate $i$ in the support of $w^t$, $w^{t+1}_{i}$ is obtained by optimizing the $i$-th coordinate (with others fixed) through:
\begin{equation}
\begin{aligned}
      w_{i}^{t+1}= \textsf{CDUpdate}(w^t,i) := &\argmin_{w\in \mathbb{R}} Q\left(w_1^{t}, \ldots, w_{i-1}^t, w, w_{i+1}^t, \ldots, w_p^t\right).
\end{aligned}
\end{equation}
Calculating $\textsf{CDUpdate}(w^t,i)$ requires the minimization of a univariate quadratic function with time cost $O(n)$.

Cyclic CD enjoys a fast convergence rate \cite{bertsekas1997nonlinear,nesterov2012efficiency}. However, the quality of the resulting solution is limited and depends heavily on the initial solution, as CD cannot modify the support of a solution. In practice, we adopt a hybrid updating rule that combines IHT and cyclic CD for better performance in terms of both quality and efficiency. In each iteration, we perform several rounds of IHT updates and then apply cyclic CD to refine the solution on the support. This approach is summarized in Algorithm \ref{alg:ihtcd}.

\begin{algorithm}[h]
\begin{algorithmic}[1]
\REQUIRE $w^0$,  $k$, $t_{\textsf{HT}},t_{\textsf{CD}}$
\FOR {$t=0,1,\ldots$}
\STATE $w\gets w^t$
\FOR {$t'=1,\ldots,t_{\textsf{HT}}$}
\STATE Compute stepsize $\tau^s$ using Algorithm \ref{alg:stepsize}.
\STATE $w\gets \textsf{HT}(w,k,\tau^s)$ \COMMENT{Time complexity $O(np)$}
\ENDFOR
\FOR {$t'=1,\ldots, t_{\textsf{CD}}$}
\FOR {$i\in\supp(w)$}
\STATE $w\gets \textsf{CDUpdate}(w,i)$ \COMMENT{Time complexity $O(n)$}
\ENDFOR
\ENDFOR
\STATE $w^{t+1}\gets w$
\ENDFOR
\end{algorithmic}
\caption{IHT with CD: \textsf{IHT-CD}$(w^0,k,t_{\textsf{HT}},t_{\textsf{CD}})$}
\label{alg:ihtcd}
\end{algorithm}

\subsection{Active set updates}\label{subsec:active-set}

The active set strategy is a popular approach that has been shown to be effective in reducing complexity in various contexts~\cite{nocedal1999numerical,friedman2010regularization,hazimeh2020fast}. In our problem setting, the active set strategy works by starting with an initial active set (of length equal to a multiple of the required number of nonzeros $k$, e.g., $2k$) that is selected based on the magnitude of the initial solution. In each iteration, we restrict the updates of Algorithm \ref{alg:ihtcd} to the current active set $\mathcal{A}$. After convergence, we perform IHT updates on the full vector to find a better solution $w$ with $\supp(w) \not\subseteq \mathcal{A}$. The algorithm terminates if such $w$ does not exist; otherwise, we update $\mathcal{A}\gets \mathcal{A}\cup \supp(w)$, and the process is repeated. Algorithm \ref{alg:active} gives a detailed illustration of the active set method, with Algorithm~\ref{alg:ihtcd} as the inner solver (potentially the inner solver can be replaced with any other solver, such as Algorithm~\ref{alg:bs} in the next section). In our experiments, this strategy works well on medium-sized problems $(p\sim10^5)$ and sparse problems $(k\ll p)$.

\begin{algorithm}[h]
\begin{algorithmic}[1]
\REQUIRE $w^0$, $k$, $t_{\textsf{HT}},t_{\textsf{CD}}$, and an initial active set $\mathcal{A}^0$
\FOR {$t=0,1,\ldots$}
\STATE $w^{t+1/2}|_{\mathcal{A}^t}\gets \textsf{IHT-CD}(w^t|_{\mathcal{A}^t},k,t_{\textsf{HT}},t_{\textsf{CD}})$ restricted on $\mathcal{A}^t$ 
\STATE Find $\tau^s$ via line search such that $w^{t+1}\gets \textsf{HT}(w^{t+1/2},k,\tau^s)$ satisfies\\ \qquad(i)~$Q(w^{t+1})<Q(w^{t+1/2})$ ~(ii)~ $\supp(w^{t+1})\not\subseteq  \mathcal{A}^t$ 
\IF {such $\tau^s$ does not exist}
\STATE \textbf{break}
\ELSE 
\STATE $\mathcal{A}^{t+1}\gets \mathcal{A}^t\cup \supp(w^{t+1})$ 
\ENDIF
\ENDFOR
\end{algorithmic}
\caption{Active set with IHT: \texttt{CHITA-CD}$(w^0,k,t_{\textsf{HT}},t_{\textsf{CD}},\mathcal{A}^0)$}
\label{alg:active}
\end{algorithm}

\subsection{Backsolve via Woodbury formula}\label{subsec:backsolve}
As the dimensionality of the problem increases, \texttt{CHITA-CD} becomes increasingly computationally expensive. To address this issue, we propose a backsolve approach that further reduces the complexity while maintaining a slightly suboptimal solution. The backsolve approach calculates the optimal solution exactly on a restricted set. We first apply IHT updates a few times to obtain an initial feasible solution $w$, and then restrict the problem to the set $\mathcal{S} := \text{supp}(w)$. Under the restriction, problem \eqref{eqn:main-problem} reduces to a quadratic problem without $\ell_0$ constraint and its minimizer reads
\begin{equation}\label{eq:bs}
    w^*_{\mathcal{S}}= (n\lambda I_k+A_{\mathcal{S}}^\top A_{\mathcal{S}})^{-1}(n\lambda \bar w_{\mathcal{S}} + A^\top_{\mathcal{S}}b),
\end{equation}
where $A_\mathcal{S}\in \mathbb{R}^{n\times k}$ denotes a submatrix of $A$ with columns only in $\mathcal{S}$. By exploiting the low-rank structure of $A_{\mathcal{S}}^\top A_{\mathcal{S}}$ and utilizing  Woodbury formula~\cite{max1950inverting}, \eqref{eq:bs} can be computed in $O(n^2k)$ operations. Specifically, one can compute~\eqref{eq:bs} using matrix-vector multiplications involving only $A_{\mathcal{S}}$ (or its transpose) and one matrix-matrix multiplication via
\begin{align}
    w_{\mathcal{S}}^*&=(n\lambda)^{-1}[ I_k-A_{\mathcal{S}}^\top(n\lambda I_k+A_{\mathcal{S}}A_{\mathcal{S}}^\top)^{-1}A_{\mathcal{S}}]\cdot(n\lambda \bar w_{\mathcal{S}}+A_{\mathcal{S}}^\top b)\nonumber\\
    &=\bar w_{\mathcal{S}}+(n\lambda)^{-1}A_{\mathcal{S}}^\top b-(n\lambda)^{-1}A_{\mathcal{S}}^\top(n\lambda I_k+A_{\mathcal{S}}A_{\mathcal{S}}^\top)^{-1}A_{\mathcal{S}}(n\lambda \bar w_{\mathcal{S}}+A_{\mathcal{S}}^\top b). \label{eqn:bs-detail}
\end{align}

The backsolve method is stated in Algorithm \ref{alg:bs}.

\begin{algorithm}[h]
\begin{algorithmic}[1]
\REQUIRE $w^0$, $k$, $t_{\textsf{HT}}$.
\STATE Construct an initial active set $\mathcal{A}^0$
\STATE $w\gets \texttt{CHITA-CD}(w^0,k,t_{\textsf{HT}},0,\mathcal{A}^0)$
\STATE $\mathcal{S}\gets \supp(w)$
\STATE $w_{\mathcal{S}}\gets (n\lambda I_k+A_{\mathcal{S}}^\top A_{\mathcal{S}})^{-1}(n\lambda \bar w_{\mathcal{S}} + A^\top_{\mathcal{S}}b_{\mathcal{S}})$, using~\eqref{eqn:bs-detail} \COMMENT{Time complexity $O(n^2k)$}
\end{algorithmic}
\caption{Backsolve: \texttt{CHITA-BSO}$(w^0,k,t_{\textsf{HT}})$}
\label{alg:bs}
\end{algorithm}

We note that prior works \cite{singh2020woodfisher,yu2022combinatorial,hassibi1992second} also use the formula, but they do not exploit the problem structure to reduce the runtime and memory consumption.

\subsection{Stratified block-wise approximation}\label{subsec:stratified-block}

We describe in this subsection a block approximation strategy whereby we only consider limited-size blocks on the diagonal of the Hessian matrix and ignore off-diagonal parts. Given a disjoint partition $\{B_i\}_{i=1}^c$ of $\{1,2,\dots,p\}$ and assume blocks of size $B_1\times B_1,\dots,B_c\times B_c$ along the diagonal, problem \eqref{eqn:main-problem} can then be decomposed into the following subproblems ($1\le i\le c$) 
\begin{equation}\label{eq:block}
    \min_{w\in\mathbb{R}^{|B_i|}} \frac12\|b_i- A_{B_i}w\|^2+\frac{n\lambda}{2}\|w-\bar w_{B_i}\|^2,~~\mathrm{s.t.}~~\|w\|_0\leq k_i,
\end{equation}
where $b_i=A_{B_i}\bar w_{B_i}-e$ and $\sum_{i=1}^ck_i=k$ determines the sparsity in each block. The difference in the selection of $\{k_i\}_{i=1}^c$ will greatly affect the quality of the solution.  We observe in experiments that the best selection strategy is to first apply magnitude pruning (or other efficient heuristics) to get a feasible solution $w$, and then set $k_i= |\supp(w)\cap B_i|,\,\forall 1\le i\le c$. Algorithm \ref{alg:block} states the block-wise approximation algorithm, with Algorithm~\ref{alg:bs} as the inner solver for each subproblem. 

In our experiment, we adopt the same strategy to employ the block-wise approximation as in the prior work \cite{yu2022combinatorial,singh2020woodfisher}. We regard the set of variables that corresponds to a single layer in the network  as a block and then subdivide these blocks uniformly such that the size of each block does not exceed a given parameter $B_{size}=10^4$. 

We clarify that the introduction of block-wise approximation is for the sake of solution quality (accuracy of pruned network) rather than algorithmic efficiency. This differs from previous works \cite{singh2020woodfisher,yu2022combinatorial}. In fact, solving \eqref{eq:block} for $i=1,\dots,c$ requires operations of the same order as solving \eqref{eqn:main-problem} directly. On the other side, we observe in our experiments that adopting block-wise approximation will dramatically increase the network MobileNet's accuracy (from 0.2\% to near 30\%, given a sparsity level of 0.8).



\begin{algorithm}[h]
\begin{algorithmic}[1]
\REQUIRE $w^{0}$, $k$, $t_{\textsf{HT}}$, a disjoint partition $\{B_i\}_{i=1}^c$ of $\{1,2,\dots,p\}$.
\STATE Obtain a feasible solution via magnitude pruning $w\gets P_k(w^0)$.
\FOR {$i=1,2,\dots,c$}
\STATE Determine sparsity level $k_i = |\supp(w)\cap B_i|$
\STATE $w_{B_i}\gets \texttt{CHITA-BSO}(w_{B_i},k_i,t_{\textsf{HT}})$
\ENDFOR
\end{algorithmic}
\caption{\texttt{CHIAT-BSO} with block approximation}
\label{alg:block}
\end{algorithm}

\section{Experiment details}\label{app:expt-detail}

\subsection{Experimental setup}\label{subsec:expt-setup-reprod}
\subsubsection{One-shot pruning experiments}\label{subsecsec:expt-one-shot}

All experiments were carried on a computing cluster. Experiments for MLPNet and ResNet20 were run on an Intel Xeon Platinum 8260 machine with 20 CPUs; experiments for MobileNetV1 and ResNet50 were run on an Intel Xeon Gold 6248 machine with 40 CPUs and one GPU. 


\paragraph{Algorithmic setting}
We utilize the \texttt{CHITA} algorithm with active set strategy and coordinate descent as acceleration techniques, as outlined in Algorithm \ref{alg:active}, to prune MLPNet and ResNet20 networks. Additionally, we use Algorithm \ref{alg:active} as the inner solver of our proposed multi-stage approach, \texttt{CHITA++}, for these networks. 
As to MobileNetV1 and ResNet50, we utilize \texttt{CHITA-BSO} with block approximation (Algorithm \ref{alg:block}) for solving single-stage problems. We employ the exact block-wise approximation strategy as applied in previous work \cite{yu2022combinatorial,singh2020woodfisher}, see Section \ref{subsec:stratified-block} for details. We also use Algorithm \ref{alg:block}  as the inner solver of our proposed multi-stage procedure \texttt{CHITA++} for these networks. In our experiments, we set the number of stages in \texttt{CHITA++} to 15 for MLPNet and ResNet20 and 100 for MobileNetV1. \texttt{CHITA++} results on MobileNetV1 are averaged over 4 runs.

\paragraph{Hyper-parameters}
For each network and each sparsity level, we run our proposed methods \texttt{CHITA} (single-stage) and \texttt{CHITA++} (multi-stage) with ridge value $\lambda$ ranging from $[10^{-5},10^3]$ and the number of IHT iterations (if Algorithm \ref{alg:active} is applied) ranging from $[5,500]$. In single-stage settings, we consider solving problem \eqref{eqn:main-problem} with/without the first-order term. We report in Table \ref{tab:accuracy} the best model accuracy over all possible hyper-parameter combinations.

\paragraph{Hyper-parameters for ResNet50 experiments} To obtain consistent results, we run \modelname~and M-FAC with the same set of hyperparameters ($\lambda = 10^{-5}, n=500, B_{size}=10^4$) and on the same training samples for Hessian and gradient approximation. We performed a sensitivity analysis with different block sizes $B_{size}$ and found similar results --- suggesting that the results are robust to the choice of $B_{size}$.

\paragraph{Fisher sample size and mini-batch size}
In practice, we replace each $\nabla \ell_i(\bar w)$ used in Hessian and gradient approximation by the average gradient of a mini-batch of size $m$. We display in Table \ref{tab:setting} the Fisher sample size $n$ and the mini-batch size $m$  (also called fisher batch size) used for gradient evaluation. Note that WoodFisher, CBS, and \texttt{CHITA} utilize the same amount of data samples and the same  batch size for MLPNet, ResNet20, and ResNet50; while for MobileNetV1, \texttt{CHITA} performs gradient evaluations on $16{,}000$ training samples, which is much less compared to WF and CBS as they require $960{,}000$ samples.

\begin{table}[h]
\centering
\begin{tabular}{c|cc|cc|cc|cc|} 
\toprule
\multirow{2}{*}{Model}  & \multicolumn{2}{c|}{MLPNet} & \multicolumn{2}{c|}{ResNet20} & \multicolumn{2}{c|}{MobileNetV1} & \multicolumn{2}{c|}{ResNet50} \\  \cmidrule{2-9} 
 &    sample     &    batch      &          sample      &    batch  &        sample     &    batch &        sample     &    batch          \\\midrule
 \texttt{CHITA} &      1000     &   1       &       1000    &   1       &      1000     &     16 & 500 & 16 \\
 WF  \& CBS &   1000        &      1    &     1000      &    1      &    400       &      2400 & - & -  \\
  M-FAC &   -        &      -   &     -      &    -      &    -       &      - & 500 & 16  \\
\bottomrule 
\end{tabular}
\vspace{10pt}
\caption{Comparisons of the Fisher sample size $n$ and the mini-batch size $m$ used in Hessian and gradient approximation on MLPNet, ResNet20, MobileNetV1 and ResNet50.  }
\label{tab:setting}
\end{table}

\subsubsection{Gradual pruning}\label{subsubsec:expt-setup-gradual pruning}

All experiments were carried on a computing cluster. Experiments for MobileNetV1 were run on an Intel Xeon Platinum 6248 machine with 30 CPUs and 2 GPUs; experiments for ResNet50 were run on five Intel Xeon Platinum 6248 machines with 200 CPUs and 10 GPUs. 

\paragraph{Details on the pruning step}
In all our gradual pruning experiments, we begin by pruning the networks to a sparsity level of 50\% and proceed with six additional pruning steps to reach the target sparsity. We follow the polynomial schedule introduced by \citet{ZhuG18} as the pruning schedule and use the \texttt{CHITA-BSO} algorithm with block approximation (Algorithm \ref{alg:block}) as the pruning method. The block size is set to $B_{size}=2000$ for MobileNetV1 and $B_{size}=500$ for ResNet50.

\paragraph{Details on the fine-tuning process}
We incorporate SGD with a momentum of $0.9$ for 12 epochs between pruning steps. Once the networks have been pruned to the target sparsity, we continue to fine-tune the networks for an additional 28 epochs using SGD with a momentum of $0.9$ (total of $100$ epochs). We utilize distributed training and set the batch size to $256$ per GPU during the SGD training process.

\begin{figure}
    \centering
\includegraphics[width=0.7\textwidth]{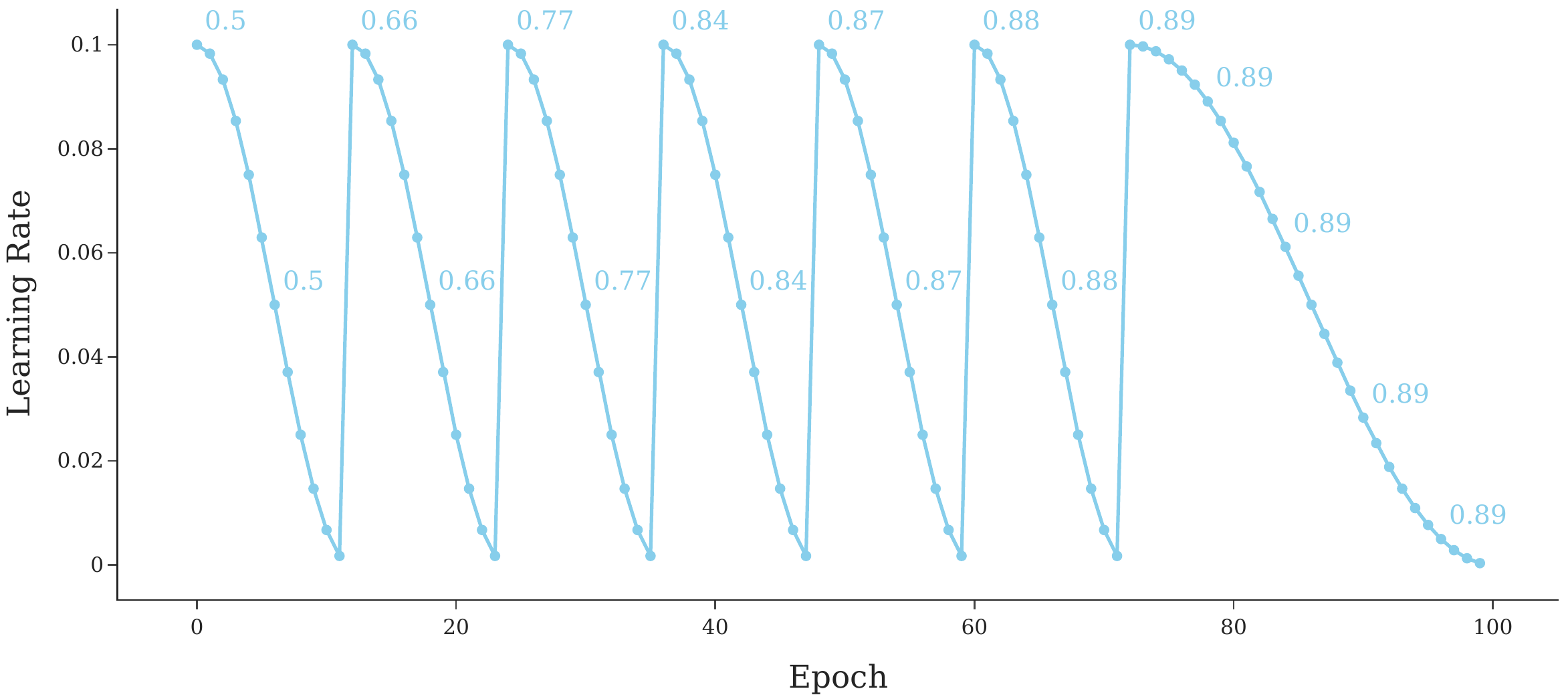}
    \caption{Learning rate schedule used in our gradual pruning experiments on MobileNetV1, with a target sparsity level of $0.89$. Text around a point indicates the sparsity of the network at the current epoch.}
    \label{fig:lr_schedule}
\end{figure}

We implement a cosine-based learning rate schedule similar to the one used in the STR method \citep{Kusupati2020STR}. Specifically, the learning rate for each epoch $e$ between two pruning steps that occur at epochs $e_1$ and $e_2$ is defined as:
 \begin{equation}
     0.00001 + 0.5 \times (0.1 - 0.00001) \times \left (1 + \cos\left(\pi\frac{ e - e_1}{e_2 - e_1} \right)\right)
 \end{equation}
 Figure \ref{fig:lr_schedule} illustrates how such a learning rate schedule decays between pruning steps.

\subsection{Implementation details and ablation studies}\label{subsec:impl-ablation}

\subsubsection{Effect of the Ridge term}\label{subsubsec:ridge}
\begin{figure*}[!h]
    \centering
        \subfigure[\label{fig:ridge1} Test accuracy with respect to the number of iterations in a single run of \texttt{CHITA-CD} (Algorithm \ref{alg:active}). The sparsity level is 0.98.]{\includegraphics[width=0.3\textwidth]{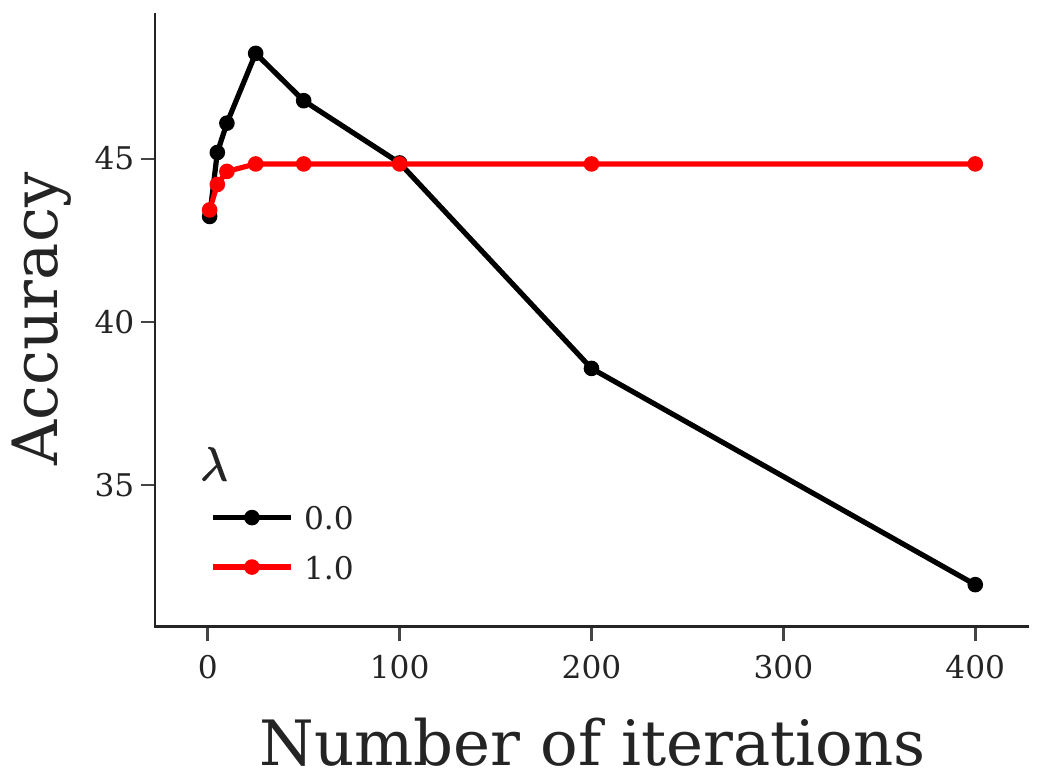}} \hfill 
     \subfigure[\label{fig:ridge2} The distance $|| w - \bar{w}||$  with respect to  the number of iterations in a single  run of \texttt{CHITA-CD} (Algorithm \ref{alg:active}). The sparsity level is 0.98.]{
\includegraphics[width=0.3\textwidth]{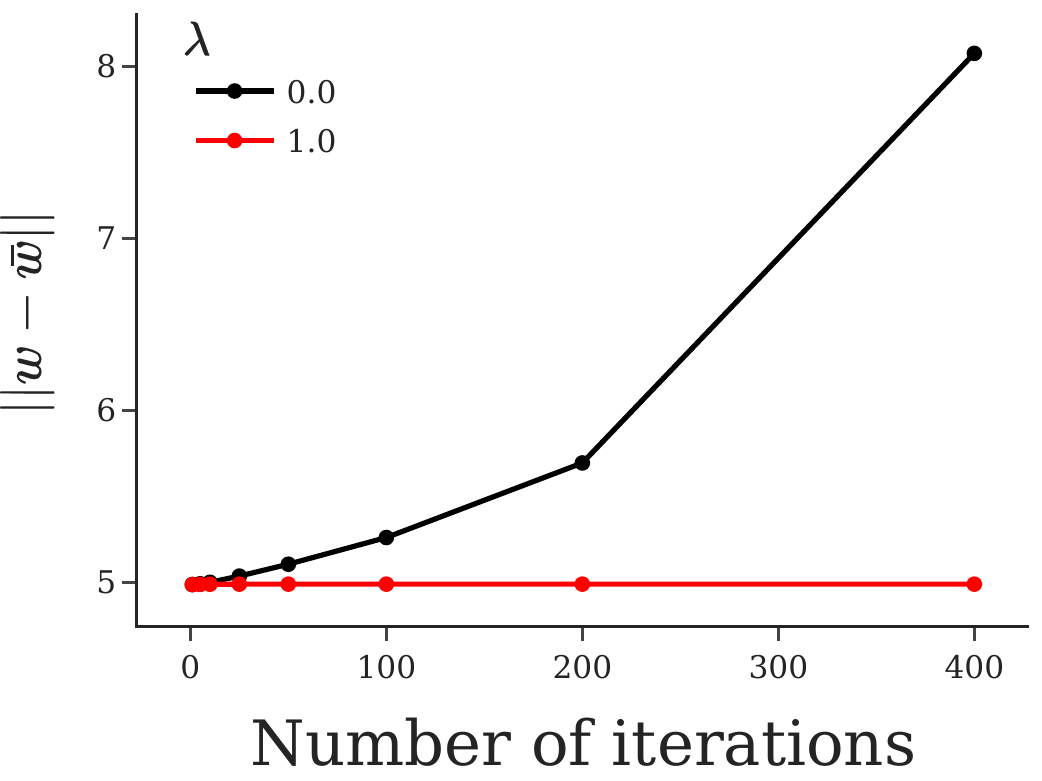}}\hfill
\subfigure[Test accuracy at different sparsity levels, with best ridge term selected in range $\lbrack  10^{-5},1 \rbrack$. The result is averaged over 4 runs. \label{fig:ridge3}]{\includegraphics[width=0.3\textwidth]{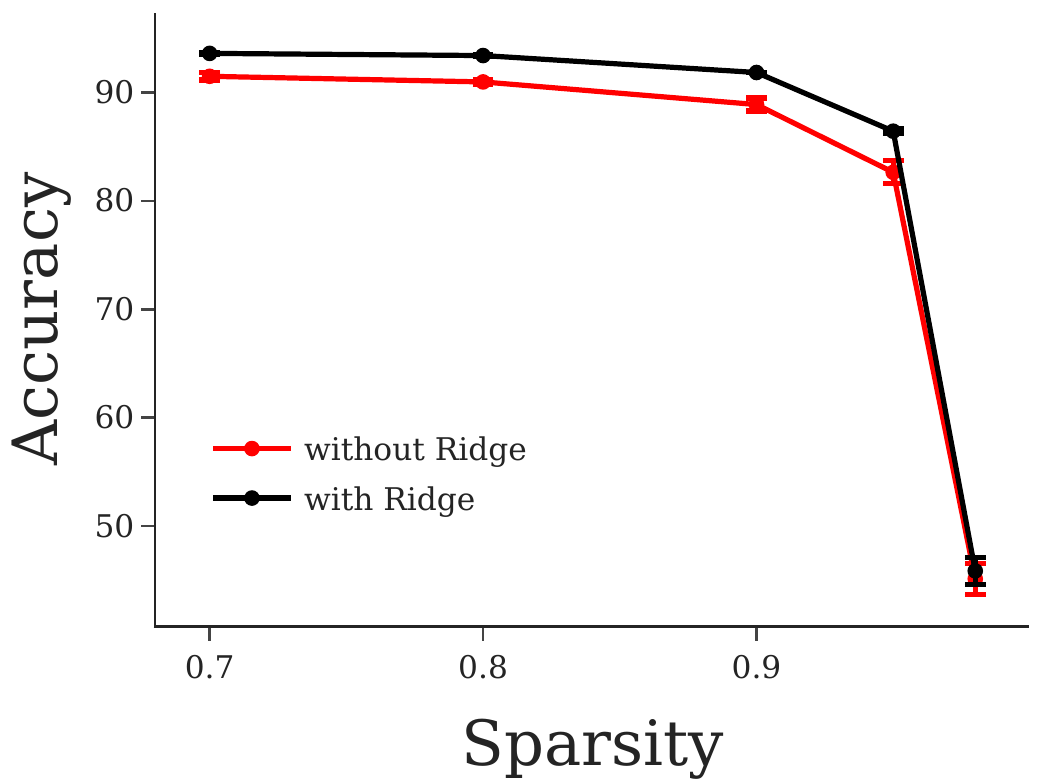}} 
\caption{Effect of the ridge term on the test accuracy and pruned weights of MLPNet.  }
     \label{fig:ridge}
\end{figure*}

In this section, we study the effect of the ridge term on the performance of our algorithm, specifically focusing on the test accuracy over the course of the algorithm. As depicted in Figure \ref{fig:ridge1}, when no ridge term is applied, the test accuracy increases initially but then experiences a sharp decline as the algorithm progresses. The underlying cause is revealed in Figure \ref{fig:ridge2}, which illustrates that without the ridge term, the distance between the original weight $\bar w$ and the pruned weight $w$ keeps increasing as the algorithm progresses. As this distance increases, the local quadratic model used in \eqref{eqn:main-problem} becomes less accurate, leading to poor test performance. 

One solution to this problem would be to employ early stopping to prevent the distance from growing too large. However, determining the optimal stopping point can be challenging. Practically, we instead add the ridge term $\frac{n\lambda}{2}\|w-\bar w\|^2$ to the objective function, effectively regularizing the model and maintaining its accuracy. As shown in Figure \ref{fig:ridge3}, utilizing a well-tuned ridge term results in an increase of approximately 3\% on MLPNet.

\subsubsection{Effect of the first-order term}\label{subsubsec:first-order}

\begin{figure*}[h]
     \centering
      \subfigure[Test accuracy with respect to mini-batch size, using different scaling factors.]{
         \includegraphics[width=0.3\textwidth]{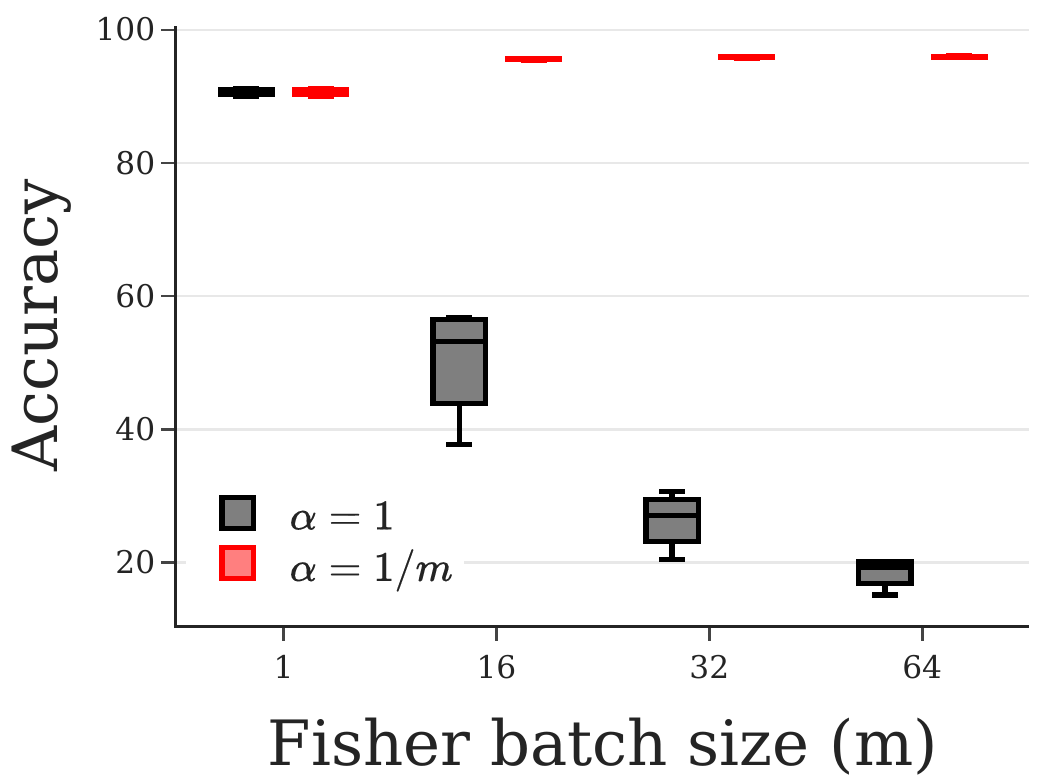}
         \label{fig:acc_alpha_1m} }
         \hfill
     \subfigure[The difference between $\alpha$ calculated by \eqref{eq:estalpha} and our proposed heuristic $1/m$.]{
         \includegraphics[width=0.3\textwidth]{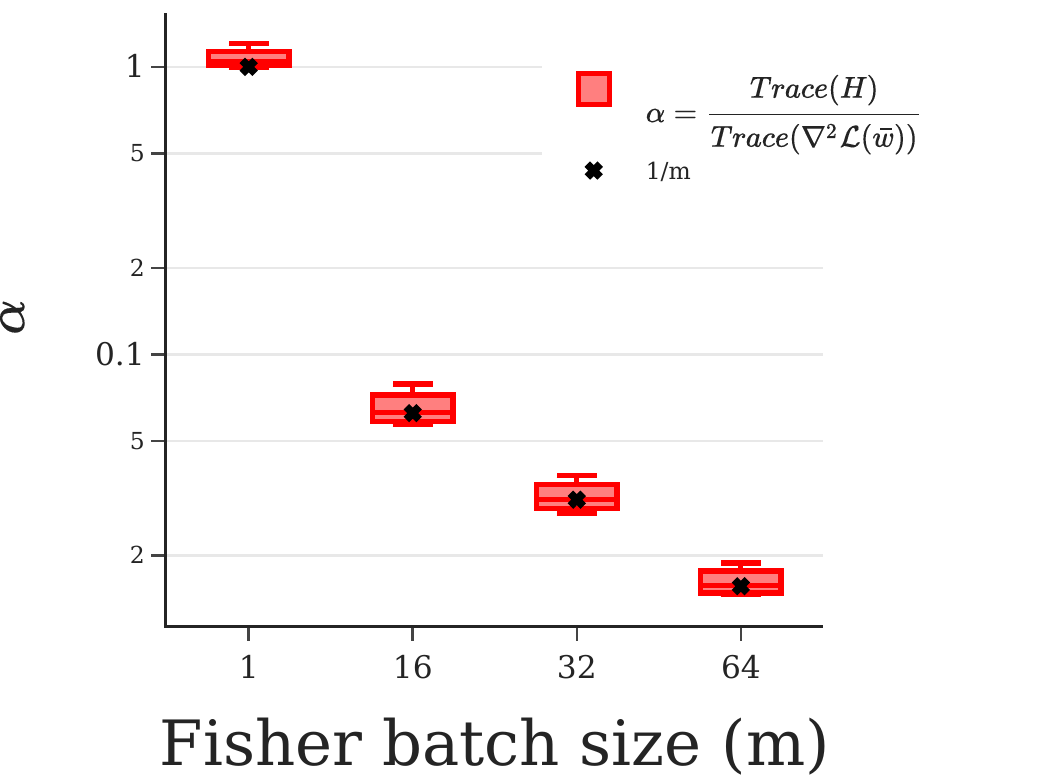}
         \label{fig:alpha}   
     }    
     \hfill
      \subfigure[Test accuracy with respect to mini-batch size, with and without the first-order term]{
         \includegraphics[width=0.3\textwidth]{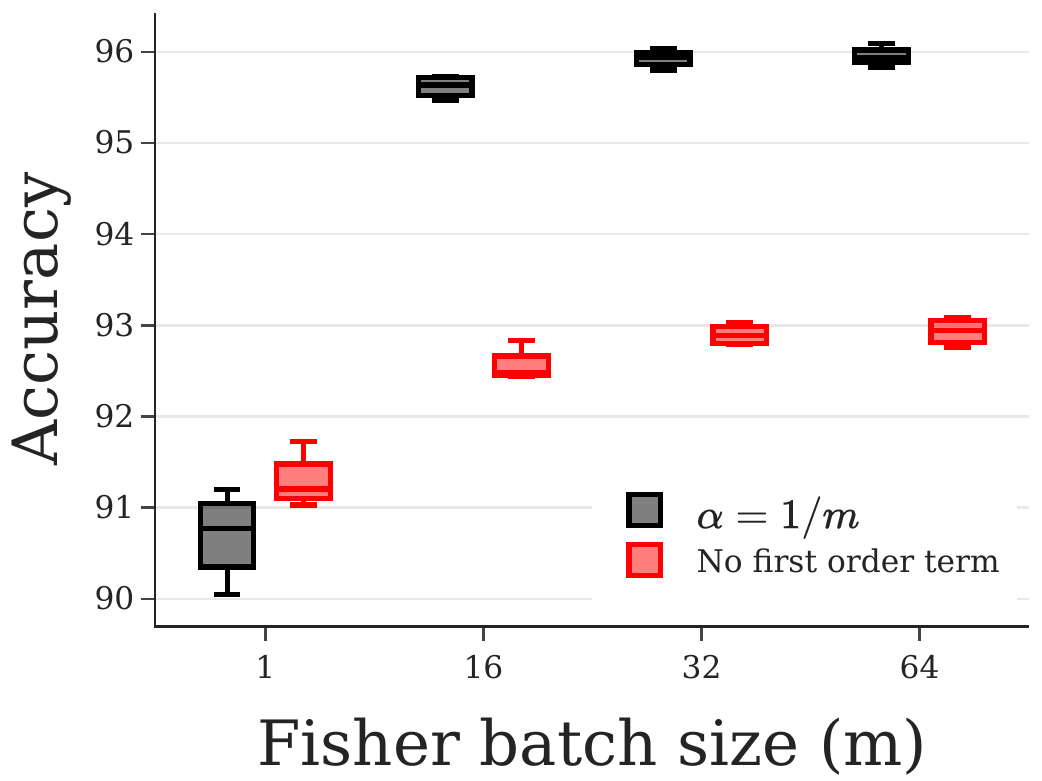}
         \label{fig:acc_alpha_1m_0}
         
     }
     \caption{Effect of using a scaled first-order term on pruning MLPNet with our proposed multi-stage solver \texttt{CHITA++} to a sparsity level of 0.95. All results are averaged over 5 runs.}
     \label{fig:first_order_term}
\end{figure*}

Mini-batches are used for gradient evaluations in practice instead of evaluating gradient ${\nabla \ell_i(\bar w)}_{i=1}^n$ on $n$ training samples. This means that each $\nabla \ell_i(\bar w)$ is replaced by the average gradient of a mini-batch of size $m$. In this scenario, the empirical Fisher matrix $H$ is not an accurate representation of the true Hessian matrix. However, it still provides a reasonable approximation but with a scaling factor~\cite{thomas2020interplay,singh2020woodfisher}.

In scale-independent applications, e.g., minimizing $\mathcal{L}(w) \approx \mathcal{L}(\bar w) + \frac12(w-\bar w)^\top H(w-\bar w)$ as considered in \citet{singh2020woodfisher} and~\citet{yu2022combinatorial},  the empirical Fisher matrix $H$ still effectively approximates the true Hessian. However, this approximation is no longer accurate in our framework, which includes a first-order term. This is supported by the results shown in Figure \ref{fig:acc_alpha_1m}, where our framework with a correctly scaled term ($\alpha=1/m$) demonstrates significantly improved performance compared to one without a scaling factor ($\alpha=1$), especially when the fisher batch size $m$ is much greater than one.

To address this issue, we propose a local quadratic approximation with a scaled first-order term that reads
\begin{equation}\label{eqn:local-quadratic3}
    Q(w)=\mathcal{L}(\bar w)+ \alpha g^\top(w-\bar w)+\frac12(w-\bar w)^\top H(w-\bar w).
\end{equation}
Our proposed $\ell_0$-constrained framework \eqref{eqn:main-problem} can be generalized to solving this problem by setting $y=A\bar w-\alpha e$, where $e$ is a vector of ones. We propose an accurate estimation of $\alpha$ as
\begin{equation}\label{eq:estalpha}
    \alpha = \frac{\text{Trace}(H)}{\text{Trace}(\nabla^2\mathcal{L}(\bar w))}.
\end{equation}
However, the computation cost of $\text{Trace}(\nabla^2\mathcal{L}(\bar w))$ is not negligible, even using accelerated methods as proposed in \citet{pyhessian}. Through experimentation, as shown in Figure \ref{fig:alpha}, we have discovered that the estimated value of $\alpha$ as given by \eqref{eq:estalpha} is relatively close to $1/m$. Therefore, we have chosen to use $1/m$ as a heuristic scaling factor in our experiments, as it provides a good approximation while reducing the computational cost.

In Figure \ref{fig:acc_alpha_1m_0}, we further illustrate the benefits of using large mini-batches and a scaled first-order term. As the fisher batch size $m$ increases, we can construct more precise local quadratic approximations through better estimation of $H$ and $g$, resulting in improved test accuracy. Additionally, when $m$ is greater than 1, using a correctly scaled first-order term provides an additional performance boost.

\end{document}